\documentclass[12pt,onecolumn,draftclsnofoot]{IEEEtran}

\usepackage{graphicx,times,psfig,amsmath} 
\usepackage{epsfig}

\begin{document}

\newtheorem{definition}{Definition}[section]
\newtheorem{lemma}{Lemma}[section]
\newtheorem{theorem}{Theorem}[section]
\newtheorem{property}{Property}[section]

\title{Cooperative Optimization for Energy Minimization: \\A Case Study of Stereo Matching}

\author{\authorblockN{Xiaofei Huang}\\
\authorblockA{School of Information Science and Technology\\
Tsinghua University, Beijing, P.~R.~China, 100084 \\
huangxiaofei@ieee.org}}
%


%


\maketitle

\begin{abstract}
Often times, individuals working together as a team
	can solve hard problems beyond the capability of any individual in the team.
Cooperative optimization is a newly proposed general method for attacking hard optimization problems
	inspired by cooperation principles in team playing.
It has an established theoretical foundation 
	and has demonstrated outstanding performances in solving real-world optimization problems. 
With some general settings, a cooperative optimization algorithm has a unique equilibrium 
	and converges to it with an exponential rate regardless initial conditions 
	and insensitive to perturbations. 
It also possesses a number of global optimality conditions
	for identifying global optima so that it can terminate its search process efficiently.
This paper offers a general description of cooperative optimization, 
	addresses a number of design issues, and presents a case study to demonstrate its power.
\end{abstract}

\section{Introduction}

Optimization is a core problem both in mathematics and computer science.
It is a very active research area 
	with many international conferences every year, a large amount of literature, 
	and many researchers and users across many fields for a wide range of applications. 
Combinatorial optimization~\cite{CPapadimitriou98,Pardalos02} is a branch of optimization
	where the set of feasible solutions of problems is discrete, countable, and of a finite size.
The general methods for combinatorial optimization are
   1) local search~\cite{Michalewicz02}, 
   2) simulated annealing~\cite{Kirkpatrick83,Geman84}, 3) genetic algorithms~\cite{Fogel66,Holland92,SchwefelES}, 
   5) ant colony optimization~\cite{Dorigo2004},
   4) tabu search~\cite{Glover97}, 5) branch-and-bound~\cite{Lawler66,Coffman76} 6) dynamic programming~\cite{Coffman76}.
The successful applications of different combinatorial optimization methods 
	have been reported in solving a large variety of optimization problems in practice.

Optimization is important in the areas of computer vision, pattern recognition, and image processing.
For example, stereo matching is one of the most active research problems in computer vision \cite{Scharstein2002,Boykov2001,Lawrence2000,Marr76}.
The goal of stereo matching is to recover the depth image of a scene from a pair of 2-D images of the same scene
   taken from two different locations.
Like many other problems from these areas, 
   it can be formulated as a combinatorial optimization problem,
   which is NP-hard \cite{Atkinson} in computational complexity in general.

The researchers in computer vision have developed 
   a number of search techniques 
   which have been proven effective in practice for finding good solutions for combinatorial optimization problems.
Two well-known ones are the cooperative algorithm proposed by D.~Marr and T.~Poggio in \cite{Marr76} for stereo matching
   and the probabilistic relaxation proposed by A.~Rosenfield et al~\cite{Rosenfeld76} for scene labeling.

Recently, there are some remarkable progresses in discovering new optimization methods 
	for solving computer vision problems.
Graph cuts~\cite{Boykov2001,Boykov2004,Scharstein2002,Kolmogorov04} is 
   a powerful specialized optimization technique popular in computer vision.
It has the best known results in energy minimization
   in the two recent evaluations of stereo algorithms~\cite{Scharstein2002,Szeliski:va99},
   more powerful than the classic simulated annealing method.
However, graph cuts has a limitation in its scope because it is only applicable 
	when the energy minimization of a vision problem can be reduced into a problem 
	of finding the minimum cut in a graph~\cite{Kolmogorov04}.
	
The second optimization method is so called the sum-product algorithm~\cite{Kschischang01}, 
	a generalized belief propagation algorithm developed in AI~\cite{Pearl88}.
The sum-product algorithm is the most powerful optimization method ever found so far
	for attacking hard optimization problems raised from channel decoding in communications.
The min-sum algorithm and max-product algorithm~\cite{Wiberg:thesis,Aji00} are its variations.
It has also been successful applied to solve several computer vision problems 
	with promising experimental results~\cite{Tappen2003}.

The third method proposed recently is so called max-product tree-reweighted message passing~\cite{Wainwright2005}.
It is based on a lower bounding technique called linear programming relaxation.
Its improvement has been proposed recently and its successful applications in computer vision have been reported~\cite{Kolmogorov2006}.

The cooperative optimization is a newly discovered general optimization method
	for attacking hard optimization problems~\cite{Huang02,Huang03Korea,Huang03Turkey,Huang03Greece}.
It has been found in the experiments~\cite{HuangBookCCO,HuangDAGM04,HuangISIT05,HuangITW06p1,HuangITW06p3} 
	that cooperative optimization has achieved remarkable performances at solving a number of real-world
	{\cal NP}-hard problems with the number of variables ranging from thousands to hundreds of thousands.
The problems span several areas, proving its generality and power.

For example, cooperative optimization algorithms have been proposed 
	for DNA image analysis~\cite{HuangBookCCO}, shape from shading~\cite{Huang03Greece}, stereo matching~\cite{Huang03Korea,HuangDAGM04}, and image segmentation~\cite{Huang04ICIP}.
In the second case, it significantly outperformed 
	the classic simulated annealing in finding global optimal solutions.
In the third case, its performance is comparable with graph cuts in terms of solution quality, 
	and is twice as faster as graph cuts in software simulation using the common evaluation framework for stereo matching~\cite{Scharstein2002}.
In the fourth case, it is ten times faster than graph cuts and has reduced the error rate by 
	two to three factors.
In all these cases, its memory usage is efficient and fixed, 
	its operations are simple, regular, and fully scalable.
All these features make it suitable for parallel hardware implementations.
	
This paper is organized in three major themes as 
	1) a formal presentation for cooperative optimization, 2) design issues, and 3) a case study.
They are the generalization and the extension of the previous papers on cooperative optimization.
In the case study, another cooperative optimization algorithm for stereo matching 
	besides the one proposed before~\cite{Huang03Korea,HuangDAGM04}
	is offered to demonstrate the power and flexibility of cooperative optimization.
Compared with the previous one for stereo matching, 
	the new one lowers the energy levels of solutions further and is more than ten times faster.
Just like the previous one, the new one is also simple in computation and fully parallel in operations, suitable for hardware implementations.

\section{Cooperative Multi-Agent System for Distributed Optimization}

Different forms of cooperative optimization can be derived from different cooperation schemes.
The basic form defines an important collection of cooperative optimization algorithms.
There are two different ways to derive it;
	namely, 1) as a cooperative multi-agent system for distributed optimization 
	and 2) as a lower bounding technique for finding global optimums.
Each way offers its own inspirations and insights to understand the algorithms.
This section describes the first way.
The following section offers the description for the second way.
Readers who are not interested in them
	can directly jump to Section~\ref{sec_gc_form} for a general description of cooperative optimization.
Those three sections are relatively independent to each other.

\subsection{Inspiration and Basic Ideas}

Team playing is a common social behavior among individuals of the same species (or different)
	where the team members working together can achieve goals or solve hard problems 
	which are beyond the capability of any member in the team. 
Often times, team playing is achieved through competition and cooperation among the members in a team. 
Usually, competition or cooperation alone can hardly lead to good solutions either for a team or 
	for the individuals in the team. 
Without competition, individuals in a team may lose motivation to pursue better solutions. 
Without cooperation, they might directly conflict with each other 
	and poor solutions might be reached both for the team and themselves. 
Through properly balanced competition and cooperation,
	individuals in a team can find the best solutions for the team and possibly good solutions for themselves at the same time.

In the terms of computer science, 
	we can view a team of this kind as a cooperative system with multiple agents. 
In the system, each agent has its own objective. 
The collection of all the agent's objectives form the objective of the system. 
We can use a cooperative system to solve a hard optimization problem 
	following the divide-and-conquer principle. 
We first break up the objective function of the optimization problem into 
	a number of sub-objective functions of manageable sizes and complexities. 
Following that, we assign each sub-objective function 
	to an agent in a system as the agent's own objective function
	and ask those agents in the system to optimize their own objective functions 
	through competition and cooperation.
(Throughout this paper, we use the term ``objective'' and ``objective function'' interchangeably
	since the objective of an optimization problem is defined by an objective function
	and this paper focuses only on optimizing objective functions.)

Specifically, the competition is achieved by asking each agent 
	to optimize its own objective function
	by applying problem-specific optimization methods or heuristics. 
However, the objectives of agents may not be always aligned with each other. 
In other words, the best solutions of the agents for optimizing their own objective functions 
	may conflict with each other. 
To resolve the conflicts, each agent passes its solution to its neighbors through local message passing. 
After receiving its neighbor's solutions, 
	each agent compromises its solution with the solutions of its neighbors. 
The solution compromising is achieved by modifying the objective function of each agent to take into account its neighbors' solutions. 
It is important to note that solution compromising among agents is a key concept 
	for understanding the cooperation strategy introduced by cooperative optimization.

Let the objective of the individual $i$ be $Objective(i)$.
Let the solution of the individual $i$ at time $t$ be $Solution(i, t)$.
Let the collection of solutions of the neighbors of the individual $i$ at time $t$ be 
	$\{Solution(j,t)|j \in Neighbors(i)\}$.
The basic operations of a cooperative system are organized as a process shown in Figure~\ref{fig_cooperative_system}.

\begin{figure}
\begin{tabbing}
123\=456\=789\=012\=345\=678\=901\ \kill 
{\em Initialization}  \\
\>For each individual $i$ in the system, find the initial solution,\\
\>\>$Find\_Solution(Objective(i)) \Rightarrow Solution(i, t = 0)$;  \\
{\em Iteration}  \\
\>For each individual $i$ in the system, \\
\>\>Modify its original objective by including its neighbors' solutions, \\
\>\>\>$Modify\_Objective(Objective(i), \{Solution(j,t)|j \in Neighbors(i)\})$ \\
\>\>\>\>$\Rightarrow Objective(i, t+1)$;\\
\>\>Find solutions of the modified objective,\\
\>\>\>$Find\_Solution(Objective(i, t+1)) \Rightarrow Solution(i, t +1)$;  
\end{tabbing}
\caption{Cooperative Multi-Agent System for Distributed Optimization.} 
\label{fig_cooperative_system}
\end{figure}

The process of a cooperative system of this kind is iterative 
	and self-organized and each agent in the system is autonomous. 
The system is also inherently distributed and parallel, 
	making the entire system highly scalable 
	and less vulnerable to perturbations and disruptions on individuals 
	than a centralized system. 
Despite of its simplicity, it has many interesting emerging behaviors 
	and can attack many challenging optimization problems.

\subsection{Basic Form of Cooperative Optimization}	
\label{section_basic_form}

In light of the cooperative multi-agent system for distributed optimization described in Fig.~\ref{fig_cooperative_system},
	we can derive the basic form of cooperative optimization now.
It is based on a direct way for defining the solution of each agent
	and a simple way to modify the objective of each agent.
The derivation can be generalized further in a straightforward way 
	to any other definitions of solutions and modifications of objectives.

Given a multivariate objective function $E(x_1, x_2, \ldots, x_n)$ of $n$ variables, 
	or simply denoted as $E(x)$, 
	where each variable $x_i$ is of a finite domain $D_i$ of size $|D_i|$. 
Assume that $E(x)$ can be decomposed into $n$ sub-objective functions $E_i(x)$, denoted as $\{E_i(x)\}$, satisfying
\begin{enumerate}
\item $E(x) = E_1(x) + E_2(x) + \ldots + E_n(x)$ \label{condition_1},
\item $E_i(x)$, for $i=1,2,\ldots,n$, contains at least variable $x_i$,
\item the minimization of $E_i(x)$, for $i=1,2,\ldots,n$, is computationally manageable in complexity.
\end{enumerate}

Let us assign $E_i(x)$ as the objective of agent $i$,
\[ Objective(i) = E_i(x), \quad \mbox{for $i=1,2,\ldots,n$} \ . \]
There are $n$ agents in the system, 
	one agent for each sub-objective function.

Let the initial solution of agent $i$ be the minimization result of $E_i(x)$ defined as follows,
\[ Solution(i, t=0) = \min_{X_i \setminus{x_i}} E_i(x) \ , \]
where $X_i$ is the set of variables contained in $E_i(x)$, 
	and $\min_{ X_i \setminus{x_i}}$ stands for minimizing with respect to all variables in $X_i$ excluding $x_i$. 
The solution is an unary function on variable $x_i$, denoted as $\Psi^{(0)}_i(x_i)$.

Assume that the system takes discrete-time with iteration step $k=1,2,3,\ldots$.
To simplify notations, let $\tilde{E}^{(k)}_i(x)$ 
	be the modified objective function of agent $i$ at iteration $k$, i.e., 
\[\tilde{E}^{(k)}_i(x)=Objective(i, t=k) \ . \]
It is also referred to as the $i$-th modified sub-objective of the system.
The agent's solution at the iteration is defined as
\begin{equation}
Solution(i, t = k) = \min_{X_i \setminus{x_i}} \tilde{E}^{(k)}_i(x)   \ . 
\label{solution_i}
\end{equation}
The solution is an unary function on variable $x_i$, denoted as $\Psi^{(k)}_i(x_i)$.
It is the state of agent $i$ at iteration $k$.
It can be represented as a vector of real values of size $|D_i|$, the domain size of variable $x_i$.
The $i$-th equation in (\ref{solution_i})
	defines the dynamics of agent $i$.
All the $n$ equations define the dynamics of the system.

As described in the previous subsection, the cooperation among the agents in the system 
	is introduced by solution compromising via modifying the objective of each agent.
Let agent $i$ define its modified objective function $\tilde{E}^{(k)}_i(x)$ at iteration $k$ as a linear combination
	of its original objective $E_i(x)$ and the solutions of its neighbors at the previous iteration $k-1$ as follows,
\begin{equation}
\tilde{E}^{(k)}_i(x) = \left(1 - \lambda_k \right) E_i(x)  + \lambda_k \sum_{j \in Neighbors(i)} w_{ij} \Psi^{(k-1)}_j(x_j) \ , 
\label{modified_objective_function}
\end{equation}
where $\lambda_k$ and $w_{ij}$ are coefficients of the linear combination.

Agent $j$ is the neighbor of agent $i$ if variable $x_j$ of the same index $j$ is contained in the agent $i$'s objective function $E_i(x)$.
(Based on this definition, the agent $i$ is also a neighbor of itself.
Such a generalization is necessary 
	because there is no restriction to have agent $i$ modify its objective using its own solution.)
The neighbors of agent $i$ is denoted as ${\cal N}(i)$, i.e., ${\cal N}(i)=Neighbors(i)$.
Specifically, it is defined as the set of indices as
\[ {\cal N}(i) = \{j|\{x_j\} \in X_i\} \ . \]

Substituting Eq.~(\ref{modified_objective_function}) into Eq.~(\ref{solution_i})
	and letting $w_{ij} = 0$ if $j \not \in {\cal N}(i)$,
	the dynamics of the cooperative system can be written as the following $n$ difference equations,
\begin{equation}
\Psi^{(k)}_i (x_i) = \min_{X_i \setminus{x_i}}\left( \left(1 - \lambda_k \right) E_i(x) + \lambda_k \sum_{j} w_{ij} \Psi^{(k-1)}_j(x_j)\right), \quad \mbox{for $i=1,2,\ldots,n$} \ . 
\label{cooperative_optimization}
\end{equation}
Such a set of difference equations defines
	a basic cooperative optimization system (algorithm) for minimizing an objective function of the form $\sum_i E_i(x)$.

At iteration $k$, variable $x_i$, for $i=1,2,\ldots,n$, has a value in the solution 
	for minimizing the $i$-th modified sub-objective function $\tilde{E}^{(k)}_i(x)$.
It is denoted as $\tilde{x}^{(k)}_i$, i.e.,
\[ \tilde{x}^{(k)}_i = \arg \min_{x_i} \min_{X_i \setminus x_i} \tilde{E}^{(k)}_i(x) \ . \]
From (\ref{solution_i}),  we have
\begin{equation}
\tilde{x}^{(k)}_i = \arg \min_{x_i} \Psi^{(k)}_i(x_i) \ . 
\label{cop_solution}
\end{equation}
The agent $i$ is responsible for assigning that value to variable $x_i$.
The assignments of other variables are taken care of by other agents.
All these values together form a solution of the system at iteration $k$, denoted as $\tilde{x}^{(k)}$.

Putting everything together, we have the pseudo code of the algorithm is given in Figure~\ref{fig_cooperative_optimization_algorithm}.
The global optimality condition mentioned in the line $7$ will be discussed 
	in detail later in this paper.
	
\begin{figure}
\begin{tabbing}
{\bf Procedure } Basic Cooperative Optimization Algorithm\\
123\=456\=789\=012\=345\=678\=901\ \kill 
1\> Initialize the soft assignment function $\Psi^{(0)}_i(x_i)$, for each $i$; \\
2\>{\bf for} $k:=1$ to $max\_iteration$ {\bf do} \\
3\>\>{\bf for} each $i$ {\bf do} \\
\>\>/* modify the $i$-th sub-objective function $E_i(x)$ */ \\
4\>\>\>$\tilde{E}^{(k)}_i(x) := (1-\lambda_k)E_i(x) + \lambda_k \sum_{j} w_{ij} \Psi^{(k-1)}_i(x_i)$ ; \\
\>\>/* minimize the modified sub-objective function */ \\
5\>\>\>$\Psi^{(k)}_i(x_i) := \min_{X_i \setminus x_i} \tilde{E}^{(k)}_i(x)$; \\
\>\>/* find the best value for $x_i$ */ \\
6\>\>\>$\tilde{x}^{(k)}_i := \arg \min_{x_i} \Psi^{(k)}_i(x_i)$; \\
7\>\>{\bf if} $\tilde{x}^{(k)}$ is a global optimal solution {\bf return} $\tilde{x}^{(k)}$; \\
8\> {\bf return} $\tilde{x}^{(k)}$; /* as an approximate solution */
\end{tabbing}
\caption{Basic cooperative optimization algorithm for minimizing an objective function of the form $E(x) = \sum^n_{i=1} E_i(x)$.}
\label{fig_cooperative_optimization_algorithm}
\end{figure}

\subsection{Cooperation Strength and Propagation Matrix}	

The coefficient $\lambda_k$ in (\ref{cooperative_optimization}) 
	controls the level of the cooperation among the agents at iteration $k$.
It is so called the cooperation strength, satisfying $0 \le \lambda_k < 1$.
From (\ref{cooperative_optimization}) we can see that, for each agent, 
	a high value for $\lambda_k$ will weigh 
        the solutions of the other agents more than its own objective $E_i(x)$.
In other words, the agents in the system tend to compromise more with their solutions.
As a consequence, 
   a strong level of cooperation is reached in this case.
If the cooperation strength $\lambda_k$ is of a small value,
	the cooperation among the agents is weak.
Particularly, if it is equal to zero,
	there is no cooperation among the agents and each agent minimizes its own objective function independently 
	(see (\ref{cooperative_optimization})).

The coefficients $w_{ij}$ control the propagation of solutions $\Psi^{(k-1)}_j(x_j)$, for $j=1,2,\ldots, n$, 
	as messages among the agents in the system.
All $w_{ij}$s together form a $n \times n$ matrix called the propagation matrix. 
To have $\sum_i E_i(x)$ as the objective function to be minimized,
	it is required~\cite{HuangBookCCO} that the propagation matrix $W=(w_{ij})_{n \times n}$ is non-negative and 
\[ \sum^n_{i=1} w_{ij} = 1, \quad \mbox{for $j=1,2,\ldots,n$} \ . \]
To have solutions $\Psi^{(k-1)}_j(x_j)$ uniformly propagated among all the agents,
	it is required~\cite{HuangBookCCO} that the propagation matrix $W$ is irreducible.
A matrix $W$ is called reducible if there exists
   a permutation matrix $P$ such that $PWP^T$
   has the block form
\[\left(
    \begin{array}{cc}
      A & B \\
      O & C 
    \end{array}
\right)\ . \]  

The role of propagation matrices in basic cooperative optimization algorithms
	is exactly same as the one of transition matrices in Markov chains (or random walks over directed graphs).
In a Markov chain, a transition matrix governs the distribution of states over time.
In a basic cooperative optimization algorithm, 
	a propagation matrix governs the distribution of solutions among agents.
The mathematical foundation for analyzing Markov chains has been well established.
They can be directly applied to analyze the message propagation of cooperative optimization.

\subsection{Soft Decisions as Messages Passed Among Agents}	
As mentioned before, the solution $\Psi^{(k)}_i(x_i)$ of agent $i$ at iteration $k$
	is an unary function on $x_i$ storing the solution of 
	minimizing the $i$-th modified sub-objective function $\tilde{E}^{(k)}_i(x)$
	(see (\ref{solution_i})).
Given a value of $x_i$, $\Psi^{(k)}_i(x_i)$ is 
	the minimal value of $\tilde{E}^{(k)}_i(x)$ with the variable $x_i$ fixed to that value.
To minimize $\tilde{E}^{(k)}_i(x)$,
	the values of $x_i$ which have smaller function values $\Psi^{(k)}_i(x_i)$
	are preferred more than those of higher function values.
The best value for assigning the variable $x_i$ 
	is the one of the minimal function value $\Psi^{(k)}_i(x_i)$ (see (\ref{cop_solution})).
Therefore, $\Psi^{(k)}_i(x_i)$ is inversely related to the preferences over different values of $x_i$
	for minimizing $\tilde{E}^{(k)}_i(x)$.
It is so called the assignment constraint on variable $x_i$, an algorithm introduced constraint on the variable.
It can also be viewed as a soft decision made by the agent for assigning the variable $x_i$ at iteration $k$.

In particular, a soft decision of agent $i$ 
	falls back to a hard decision for assigning the variable $x_i$
	when the agent accept only one value and reject all the rest values.
Such a hard decision can be represented by the assignment constraint $\Psi^{(k)}_i(x_i)$ as $\Psi^{(k)}_i(\tilde{x}_i) = 0$, for some $\tilde{x}_i \in D_i$,
	and $\Psi^{(k)}_i(x_i) = \infty$ for any $x_i \not = \tilde{x}_i$.

With that insight, it can be understood now that
	the messages propagated around among the agents in a basic cooperative optimization system
	are the soft decisions for assigning variables.
An agent can make a better decision using soft decisions propagated from its neighbors
	than using the hard ones instead.
It is important to note that soft decision making is a critical feature 
	of cooperative optimization,
	which makes it fundamentally different from many classic optimization methods
	where hard decisions are made for assigning variables.

\subsection{A Simple Example}
\label{sec_simple_example}

Given an objective function of the following form
\begin{eqnarray}
\lefteqn{E(x_1, x_2, x_3, x_4, x_5) = } \nonumber \\
&& f_1(x_1) + f_2(x_2) + f_3(x_3) + f_4(x_4) + f_5(x_5) + \nonumber\\
&& f_{1,2}(x_1, x_2) + f_{2,3}(x_2, x_3) + f_{3,4}(x_3, x_4) + \nonumber\\
&& f_{4,5}(x_4, x_5) + f_{1,5}(x_1, x_5) + f_{2,5}(x_2, x_5) \ ,
\label{obj_simple_example}
\end{eqnarray}
where each variable is of a finite domain.
The goal is to seek values (labels) of the five variables such that the objective function is minimized.

Let us simply denote the function as
\begin{eqnarray*}
\lefteqn{E(x) = f_1 + f_2 + f_3 + f_4 + f_5 + } \\
&& f_{1,2} + f_{2,3} + f_{3,4} +f_{4,5} + f_{1,5} + f_{2,5} \ .
\end{eqnarray*}

To design a basic cooperative optimization algorithm to minimize the objective function,
	we first decompose it into the following five sub-objective functions, 
\begin{eqnarray*}
\noindent
E_1(x_1, x_2, x_5)~~~~ &=& f_1 + f_{1,2}/2 + f_{1,5}/2; \\
E_2(x_1, x_2, x_3, x_5) &=& f_2 + f_{1,2}/2 + f_{2,3}/2 + f_{2,5}/2; \\
E_3(x_2, x_3, x_4)~~~~ &=& f_3 + f_{2,3}/2 + f_{3,4}/2;\\
E_4(x_3, x_4, x_5)~~~~&=& f_4 + f_{3,4}/2 + f_{4,5}/2;\\
E_5(x_1, x_2, x_4, x_5) &=& f_5+f_{1,5}/2 + f_{2,5}/2 + f_{4,5}/2. 
\end{eqnarray*}

A propagation matrix $W$ of dimensions $5 \times 5$ can be chosen as
\begin{equation}
W = \left(
   \begin{array}{ccccc}
   0 & \frac{1}{3} & 0 & 0 & \frac{1}{3} \\
   \frac{1}{2} & 0 & \frac{1}{2} & 0 & \frac{1}{3} \\
   0 & \frac{1}{3} & 0 & \frac{1}{2} & 0  \\
   0 & 0 & \frac{1}{2} & 0 & \frac{1}{3} \\
   \frac{1}{2} & \frac{1}{3} & 0 & \frac{1}{2} & 0 
   \end{array}
\right) 
\label{p2}
\end{equation}

With the decomposition and the propagation matrix, 
	substituting them into (\ref{cooperative_optimization}) 
	we have a basic cooperative optimization algorithm with five difference equations
	for minimizing the five sub-objective functions in an iterative and cooperative way.

\subsection{Basic Canonical Form as Generalization}	

Replacing $\Psi^{(k)}_i(x)$ by $(1-\lambda_k)\Psi^{(k)}_i(x)$ in the difference equations~(\ref{cooperative_optimization}),
	we have the basic canonical form of cooperative optimization as
\begin{equation}
\Psi^{(k)}_i (x_i) = \min_{X_i \setminus{x_i}}\left( E_i(x) + \lambda_k \sum_{j} w_{ij} \Psi^{(k-1)}_j(x_j)\right), \quad \mbox{for $i=1,2,\ldots,n$} \ . 
\label{cooperative_optimization_canonical}
\end{equation}

The basic form of cooperative optimization~(\ref{cooperative_optimization})
	has its cooperation strength $\lambda_k$ restricted to $0 \le \lambda_k < 1$.
It is because its difference equations~(\ref{cooperative_optimization}) do not make sense when $\lambda_k \ge 1$.
However, such a restriction can be relaxed to $0 \le \lambda_k$ 
	for the basic canonical form~(\ref{cooperative_optimization_canonical}). 
Often in times in practice, the basic canonical form is preferred over the basic one
	because the cooperation strength $\lambda_k$ in the former has a broader range to choose from 
	to maximize performance.

\section{Cooperative Optimization as Lower Bounding Technique}
\subsection{Bound Function Tightening Technique for Optimization}
\begin{figure}
\centering
\center{\epsfxsize 10.0cm \epsffile{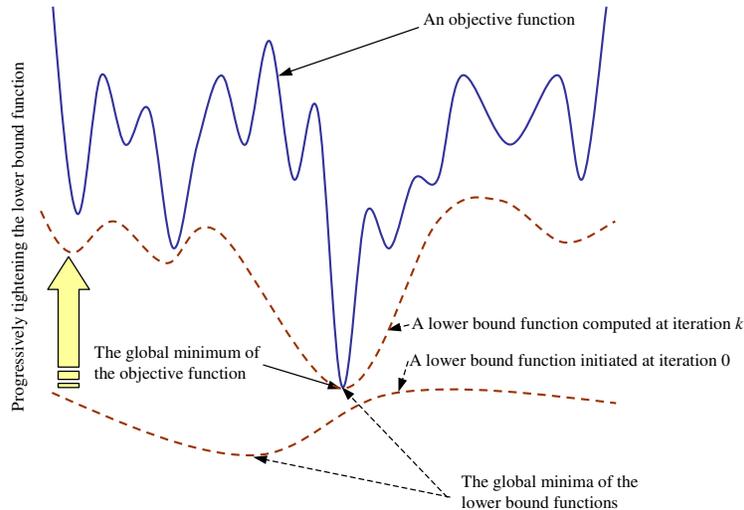}}
\caption{The global minimum of a complex multivariate objective function can be found by progressively tightening a lower bound function of some simple form
	until its global minimum touches the one of the original objective function. }
\label{fig_LowerBound}
\end{figure}

In principle, a basic cooperative optimization algorithm
	can be understood as a lower bounding technique for finding global minima.
It first initializes a function of some form 
	as a lower bound function to an objective function. 
One may intentionally choose a form for the lower bound function
	such that the minimization of the function is simple in computation.
Following that, the algorithm progressively tightens the lower bound function until its global minimum 
	touches the global minimum of the original objective function. 
The latter is then found 
	by searching the former instead (see the illustration in Fig.~\ref{fig_LowerBound}).

Specifically, let the objective function to be minimized be $E(x)$. 
Assume that the initial lower bound function be $E^{(0)}_{-}(x)$, $E^{(0)}_{-}(x) \le E(x)$.
From $E^{(0)}_{-}(x)$, assume that the algorithm progressively tightens the function in an iterative way such that
\[ E^{(0)}_{-}(x) \le E^{(1)}_{-}(x) \le \ldots \le E^{(k)}_{-}(x) \le E(x) \ , \]
where $k$ is the iteration number.

Let the global minimum of the lower bound function $E^{(k)}_{-}(x)$ at iteration $k$ be $\tilde{x}^{(k)}$.
Finding $\tilde{x}^{(k)}$ is simple in computation due to the simple form of the lower bound function $E^{(k)}_{-}(x)$.
At iteration $k$, if the algorithm found that 
	the lower bound function $E^{(k)}_{-}(x)$ at the solution $\tilde{x}^{(k)}$ 
	has the same function value as the original objective function $E(x)$, i.e.,
\[ E^{(k)}_{-}(\tilde{x}^{(k)}) = E(\tilde{x}^{(k)}) \ . \]
In other words, the two functions touch each other at the point where $x= \tilde{x}^{(k)}$ in the search space.
Then $\tilde{x}^{(k)}$ must also be the global minimum of $E(x)$ simply because 
\begin{equation}
E(\tilde{x}^{(k)}) = E^{(k)}_{-}(\tilde{x}^{(k)}) \le E^{(k)}_{-}(x) \le E(x), \quad \mbox{for any $x$} \ . 
\label{monotonically_nondecreasing}
\end{equation}
Such a condition implies that the lower bound function $E^{(k)}_{-}(x)$ has been tightened enough 
	such that its global minimum $\tilde{x}^{(k)}$ touches the global minimum of the original objective function $E(x)$.
The latter is thus found
	by searching the former instead.

Such a lower bounding technique is 
	so called the bound function tightening technique for optimization.
There are other lower bounding techniques based on principles different from this one.
Examples are Lagrangian relaxation techniques, cutting plane techniques, 
	branch-and-bound algorithms, and branch-and-cut algorithms. 

\subsection{Basic Form as Lower Bounding Technique}
In light of the bound function tightening technique described in the previous subsection,
	we can derive the basic form of cooperative optimization
	based on a simple form of lower bound functions.
The derivation can be generalized further in a straightforward way to any other forms of lower bound functions.

Given an objective function of $n$ variables, $E(x_1, x_2, \ldots, x_n)$, or simply denoted as $E(x)$. 
Assume that $E_{-} (x)$ is a lower bound function of $E(x)$ defined on the same set of variables.
Obviously the linear combination of the two functions,
\begin{equation}
(1-\lambda) E(x) + \lambda E_{-}(x) \ , 
\label{new_LBF}
\end{equation}
defines a new lower bound function of $E(x)$ if the parameter $\lambda$ satisfying $0 \le \lambda < 1$.

Let us choose a simple form for the lower bound function as
\begin{equation}
E_{-} (x) = \Psi_1(x_1) + \Psi_2(x_2) + \ldots + \Psi_n(x_n) \ , 
\label{simple_LBF}
\end{equation}
where $\Psi_i(x_i)$ is an unary component function defined on variable $x_i$, for $i=1,2,\ldots, n$.
Its global minimum, denoted as $\tilde{x}$, can be easily found  
	by minimizing the unary component functions $\Psi_i(x_i)$ independently as
\[ \tilde{x}_i = \arg \min_{x_i} \Psi_i (x_i), \quad \mbox{for $i = 1, 2, \ldots, n$} \ . \]

Assume that the objective function $E(x)$ can be decomposed into $n$ sub-objective functions,
\[ E(x) = E_1(x) + E_2(x) + \ldots + E_n(x) \ . \]
The lower bound function $E_{-} (x)$ can also be easily decomposed into $n$ sub-functions as follows
\begin{eqnarray*}
\lefteqn{E_{-}(x) = \sum^{n}_{i=1} w_{ij} \Psi_j (x_j), \quad \mbox{where }} \\
&& w_{ij} \ge 0 \mbox{ and } \sum_i w_{ij} = 1, \quad \mbox{for $1 \le i,j \le n$} \ . 
\end{eqnarray*}
Based on the two decompositions, 
	the new lower bound function~(\ref{new_LBF}) can be rewritten as
\begin{equation}
\sum^{n}_{i=1} \left( (1 - \lambda ) E_i(x) + \lambda \sum_{j} w_{ij} \Psi_j(x_j) \right) \ . 
\label{LBF1}
\end{equation}
To put the above function in a simple form, let 
\[ \tilde{E}_i(x) = (1-\lambda) E_i(x) + \lambda \sum_{j} w_{ij} \Psi_j(x_j) \ . \]
Then it can be rewritten simply as
\[ \sum^{n}_{i=1} \tilde{E}_i(x) \ .\]
In the above sum, let $\tilde{X}_i$ be the set of variables contained in the $i$-th component function $\tilde{E}_i(x)$.
If we minimize the function with respect to all variables in $\tilde{X}_i$ except for $x_i$, 
	we obtain an unary function defined on $x_i$, denoted as $\Psi^{'}_i(x_i)$, i.e.,
\begin{equation}
\Psi^{'}_i(x_i) = \min_{\tilde{X}_i \setminus x_i} \tilde{E}_i(x) \ . 
\label{compute_psi_i}
\end{equation}
The sum of those unary functions defines another lower bound function of $E(x)$, denoted as $E^{'}_{-}(x)$, i.e.,
\[ E^{'}_{-}(x) = \sum^{n}_{i=1} \Psi^{'}_i(x_i) \le E(x) \  . \]
This new lower bound function has exactly the same form as the original one $E_{-}(x) = \sum_i \Psi_i(x_i)$. 
Therefore, from a lower bound function $E_{-}(x)$ of the form $\sum_i \Psi_i(x_i)$,
	we can compute another lower bound function $E^{'}_{-}(x)$ of the same form.
Such a process can be repeated and we can have an iterative algorithm 	
	to compute new lower bound functions.

Rewriting Eq.~(\ref{compute_psi_i}) in an iterative format, we have
\begin{equation}
\Psi^{(k)}_i (x_i) = \min_{\tilde{X}_i \setminus{x_i}} \left(\left(1 - \lambda_k \right) E_i(x) + \lambda_k \sum_{j} w_{ij} \Psi^{(k-1)}_j(x_j) \right) ,
\label{cooperative_optimization2}
\end{equation}
where $k$ is the iteration step, $k=1,2,3,\ldots$.
The above $n$ difference equations define a basic cooperative optimization algorithm 
	for minimizing an objective function $E(x)$ of the form $\sum_i E_i(x)$.

The solution at iteration $k$, denoted as $\tilde{x}^{(k)}$,
	is defined as the global minimal solution of the lower bound function $E^{(k)}_{-}(x)$ at the iteration, i.e.,
\[ \tilde{x}^{(k)} = \arg \min_{x} E^{(k)}_{-}(x), \]
which can be easily obtained as
\begin{equation}
\tilde{x}^{(k)}_i = \arg \min_{x_i} \Psi^{(k)}_i(x_i), \quad \mbox{for $i=1,2,\ldots,n$} \ . 
\label{cop_solution2}
\end{equation}

If $E^{(k)}_{-}(\tilde{x}^{(k)}) = E(\tilde{x}^{(k)})$ at some iteration $k$, 
	then the solution $\tilde{x}^{(k)}$ must be the global minimum of the original objective function $E(x)$.

Without loss of generality,  we assume in the following discussions that all sub-objective functions $E_i(x)$ 
	are nonnegative ones.
One may choose the initial condition as $\Psi^{(0)}_i(x_i) = 0$, 
	for any value of $x_i$ and $i=1,2,\ldots, n$.
The parameter $\lambda_k$ can be varied from one iteration to another iteration.
If it is of a constant value and the above initial condition has been chosen,
	cooperative optimization theory~\cite{HuangBookCCO} tells us that the lower bound function $E^{(k)}_{-}(x)$
	is monotonically non-decreasing as shown in (\ref{monotonically_nondecreasing}).

\section{Computational Properties}	
	
\subsection{General Convergence Properties of Cooperative Optimization}	

It has been shown that a basic cooperative optimization algorithm~(\ref{cooperative_optimization})
	has some important computational properties~\cite{HuangBookCCO}. 
Given a constant cooperation strength $\lambda$, i.e., $\lambda_k = \lambda$ for all $k$s,
	the algorithm has one and only one equilibrium. 
It always converges to the unique equilibrium with an exponential rate 
	regardless of initial conditions and perturbations. 
The two convergence theorems proved in \cite{HuangBookCCO} are very important and so they are listed here again.
One formally describes the existence and the uniqueness of the equilibrium of the algorithm,
	and the another reveals the convergence property of the algorithm.

\begin{theorem}
A basic cooperative optimization algorithm with a constant cooperation strength $\lambda$ ($0 \le \lambda < 1$)
	has one and only one equilibrium.
That is, the difference equations~(\ref{cooperative_optimization}) of the algorithm
   have one and only one solution (equilibrium), denoted as a vector $(\Psi^{(\infty)}_1,\Psi^{(\infty)}_2,\ldots,\Psi^{(\infty)}_n)^T$, or simply $\Psi^{(\infty)}$.
\label{theorem_7}
\end{theorem}

\begin{theorem}
A basic cooperative optimization algorithm with a constant cooperation strength $\lambda$ ($0 \le \lambda < 1$)
   converges exponentially to its unique equilibrium $\Psi^{(\infty)}$ with the rate $\lambda$ 
   with any choice of the initial condition $\Psi^{(0)}$.
That is,
\begin{equation}
\|\Psi^{(k)}-\Psi^{(\infty)}\|_{\infty} \le 
   \lambda^k \|\Psi^{(0)} -\Psi^{(\infty)}\|_{\infty}\ .
\end{equation}
where $\|x\|_{\infty}$  is the maximum norm of the vector $x$ defined as
\[ \|x\|_{\infty}=\max_i|x_i| \ . \]
\label{theorem_8}
\end{theorem}

The two theorems indicate that
   every basic cooperative optimization algorithm~(\ref{cooperative_optimization}) is stable and
   has a unique attractor, $\Psi^{(\infty)}$.
Hence, the evolution of the algorithms is robust,
   insensitive to perturbations.
The final solution of the algorithms
   is independent of their initial conditions.
In contrast, the conventional algorithms
   based on iterative local improvement of solutions
   may have many local attractors due to the local minima problem.
The evolution of those local optimization algorithms are sensitive 
   to perturbations,
   and the final solution of those algorithms is dependent on their initial conditions.

Furthermore, the basic cooperative optimization algorithms~(\ref{cooperative_optimization}) 
	possess a number of global optimality conditions for identifying global optima.
They know whether a solution they found is a global optimum
	so that they can terminate their search process efficiently. 
However, this statement does not imply that {\it NP=P} because a basic cooperative optimization algorithm 
	can only verify within a polynomial time whether a solution it found is a global optimum or not. 
It cannot decide the global optimality for any given solution other than those it found. 
	
It is important to note that a basic canonical cooperative optimization algorithm~(\ref{cooperative_optimization_canonical}) 
	may no longer possess the unique equilibrium property when its cooperation strengths at some iterations
	are greater than one, i.e., $\lambda_k > 1$ for some $k$s.
In this case, the algorithm may have multiple equilibriums.
It can evolve into any one of them depending 
	on its initial settings of the assignment constraints $\Psi^{(0)}_i(x_i)$ $(1 \le i \le n)$.
	
\subsection{Consensus Solution and Solution Consensus in Distributed Optimization}	
\label{sec_global_optimality}

As described before,
	a basic cooperative optimization algorithm is defined by the $n$ difference equations~(\ref{cooperative_optimization}).
The $i$-th equation defines the minimization of
	the $i$-th modified sub-objective function $\tilde{E}^{(k)}_i(x)$ (defined in (\ref{modified_objective_function})).
Given any variable, say $x_i$, 
	it may be contained in several modified sub-objective functions. 
At each iteration, $x_i$ has a value in the optimal solution for minimizing each of the modified sub-objective functions containing the variable. 
Those values may not be the same. 
If all of them are of the same value at some iteration, 
	we say that the cooperative optimization algorithm reach a consensus assignment for that variable. 
Moreover, if a consensus assignment is reached for every variable of the problem at hand at some iteration, 
	we call the minimization of the $n$ modified sub-objective functions reaches a solution consensus.
That is, there is no conflict among the solutions in terms of variable assignments for minimizing those functions.
In this case, those consensus assignments form a solution, called a consensus solution,
	and the algorithm is called reaching a consensus solution.

To be more specific, 
	given $n$ modified sub-objective functions, $\tilde{E}_i(x)$, for $i=1,2,\ldots, n$ 
	(to simplify notation, let us drop the superscript $k$ temporarily).
Let the optimal solution of the $i$-th modified sub-objective function be $\tilde{x}(\tilde{E}_i)$, i.e.,
\[ \tilde{x}(\tilde{E}_i) = \arg \min_{x} \tilde{E}_i(x) \ . \]
Assume that variable $x_i$ is contained in both $j$-th and $k$-th modified sub-objective functions $\tilde{E}_j(x)$, $\tilde{E}_k(x)$.
However, it is not necessary that
\[ \tilde{x}_i(\tilde{E}_j) = \tilde{x}_i(\tilde{E}_k) \ . \]
Given a variable $x_i$, if the above equality holds for any $j$ and $k$ where $\tilde{E}_j(x)$ and $\tilde{E}_k(x)$ contain $x_i$,
	then a consensus assignment is reached for that variable with the assignment value denoted as $\tilde{x}_i$.
Moreover, if the above statement is true for any variable, 
	we call the minimization for all $\tilde{E}_i(x)$s reaches a solution consensus.
The solution $\tilde{x}$ with $\tilde{x}_i$ as the value of variable $x_i$ is called a consensus solution.

As defined before, $\tilde{X}_i$ stands for the set of variables contained in the function $\tilde{E}_i(x)$.
$\tilde{X}_i$ is a subset of variables, i.e.,
$\tilde{X}_i \subseteq X = \{x_1, x_2, \ldots, x_n \}$.
Let $\tilde{x}(\tilde{X}_i)$ stand for the restriction of a solution $\tilde{x}$ on $\tilde{X}_i$.
Another way to recognize a consensus solution $\tilde{x}$
	is to check if $\tilde{x}(\tilde{X}_i)$, for any $i$, is the global minimum of $\tilde{E}_i(x)$, 
	i.e.,
\[ \tilde{x}(\tilde{X}_i) = \arg \min_{x} \tilde{E}_i(x), \quad \mbox{for any $i$ } \ . \]

Simply put, a solution is a consensus one 
	if it is the global minimum of every modified sub-objective function.

\subsection{Consensus Solution in Cooperative Optimization}	

Consensus solution is an important concept of cooperative optimization.
If a consensus solution is found at some iteration or iterations,
   then we can find out the closeness between 
   the consensus solution and the global optimal solution in cost.
The following theorem from \cite{HuangBookCCO} makes these points clearer.

\begin{theorem}
Let
\[ E^{*(k)}_{-} = \sum^n_{i=1} \Psi^{(k)}_i(\tilde{x}^{(k)}_i), \quad \mbox{where }\tilde{x}^{(k)}_i = \arg \min_{x_i} \Psi^{(k)}_i(x_i) \ . \]
Given any propagation matrix $W$,
   and the general initial condition $\Psi^{(0)}_i(x_i)=0$, for each $i$, or $\lambda_1=0$.
If a consensus solution $\tilde{x}$ is found
   at iteration $k_1$ and remains the same from iteration $k_1$ to iteration $k_2$,
then the closeness between the cost of $\tilde{x}$,
    $E(\tilde{x})$, and the optimal cost, $E^{*}$, satisfies 
   the following two inequalities,
\begin{equation}
0 \le E(\tilde{x})- E^{*} \le 
   \left(\prod^{k_2}_{k=k_1} \lambda_k\right) \left(E(\tilde{x})-E^{*(k_1-1)}_{-}\right),
\end{equation}
\begin{equation}
0 \le E(\tilde{x})- E^{*} \le 
   \frac{\prod^{k_2}_{k=k_1} \lambda_k}{1-\prod^{k_2}_{k=k_1}\lambda_k} (E^{*}-E^{*(k_1-1)}_{-})\ ,
\end{equation}
where $(E^{*}-E^{*(k_1-1)}_{-})$ is the difference 
   between the optimal cost $E^{*}$ and
   the lower bound on the optimal cost $E^{*(k_1-1)}_{-}$
   obtained at iteration $k_1 - 1$.

In particular, if $1 - \lambda_k \ge \epsilon > 0$, for $k_1 \le k \le k_2$, when $k_2-k_1 \rightarrow \infty$,
\[E(\tilde{x}) \rightarrow E^{*} \ . \]
That is, the consensus solution $\tilde{x}$ must be global minimum of $E(x)$, i.e.,${\tilde x} = x^{*}$.
\label{theorem_2}
\end{theorem}  

Consensus solution is also an important concept of cooperative optimization 
	for defining global optimality conditions. 
The cooperative optimization theory tells us that a consensus solution can be the global minimal solution. 
As mentioned in the previous subsection that
	a basic cooperative optimization algorithm has one and only one equilibrium
	given a constant cooperation strength.
If a cooperative optimization algorithm reaches an equilibrium after some number of iterations 
	and a consensus solution is found at the same time, 
	then the consensus solution must be the global minimal solution, guaranteed by theory.
The following theorem (with its proof in the appendix) establishes 
	the connection between a consensus solution and a global optimal solution.

\begin{theorem}
\label{sufficient_condition}
Assume that a basic cooperative optimization~(\ref{cooperative_optimization})
	reaches its equilibrium at some iteration, denoted as $\Psi^{(\infty)}$.
That is, $\Psi^{(\infty)}$ is a solution to the difference equations~(\ref{cooperative_optimization}).
If a consensus solution ${\tilde x}$ is found at the same iteration, 
	then it must be the global minimum of $E(x)$, i.e.,${\tilde x} = x^{*}$.
\end{theorem}

Besides the basic global optimality condition given in the above theorem, 
	a few more ones are offered in \cite{HuangBookCCO} 
	for identifying global optimal solutions.
The capability of recognizing global optimums
	is a critical property for any optimization algorithm.
Without any global optimality condition,
	it will be hard for an optimization algorithm to know where to find global optimal solutions 
	and whether a solution it found is a global optimum. 
Finding ways of identifying global optimums for any optimization algorithm
	is of both practical interests as well as theoretical importance.

\subsection{Further Generalization of Convergence Properties}	
The convergence theorem~\ref{theorem_2} can be generalized further
	to any initial conditions for $\Psi^{(0)}$ and $\lambda_1$,
	and to any cooperation strength series $\{\lambda_k\}_{k \ge 1}$.
Dropping the restriction on the initial conditions $\Psi^{(0)}$ and $\lambda_1$ in the theorem,
	from the difference equations~(\ref{cooperative_optimization}), we have
\begin{equation}
E^{*} - E^{*(k_2)}_{-} = \left(\prod^{k_2}_{k=k_1} \lambda_k \right)(E^{*} - E^{*(k_1-1)}_{-}) \ . 
\label{generalized_convergence}
\end{equation}
It is obvious from the above equation that 
$E^{*(k_2)}_{-}$ still approaches $E^{*}$ exponentially with the rate $\lambda$
	when the cooperation strength $\lambda_k$ is of a constant value $\lambda$ ($0 \le \lambda < 1$). 

When the cooperation strength $\lambda_k$ is not of a constant value $\lambda$,
	the convergence to the global optimum is still guaranteed as long as 
	the cooperation strength series $\{1-\lambda_k\}_{k \ge 1}$ is divergent.

\begin{lemma}[Infinite Products]
Let $\{\lambda_k\}_{k \ge 1}$ be a sequence of numbers of the interval $[0,1)$.
\begin{enumerate}
\item If $\sum^{\infty}_{k=1} (1-\lambda_k) < \infty$, then
\[ \lim_{n \rightarrow \infty} \prod^{n}_{k=1} \lambda_k > 0 \ . \]
\item If $\sum^{\infty}_{k=1} (1-\lambda_k) = \infty$, then
\[ \lim_{n \rightarrow \infty} \prod^{n}_{k=1} \lambda_k =0 \ . \]
\end{enumerate}
\label{lemma_infinite_product}
\end{lemma}

The proof of the lemma is offered in Appendix.

From the above lemma and Eq.~(\ref{generalized_convergence}), the convergence theorem~\ref{theorem_2} can be generalized further as follows.

\begin{theorem}
Given any initial conditions,
	assume that a consensus solution $\tilde{x}$ is found by a basic cooperative optimization algorithm
   	at some iteration $k$ and remains the same in the following iterations.
If the series
\begin{equation}
(1 - \lambda_1) + (1 - \lambda_2) + \ldots + (1 - \lambda_k) + \ldots \ , 
\label{series_lambda}
\end{equation}
is divergent,
   then
\[ E(\tilde{x}) = E^{*}. \]
That is, the consensus solution $\tilde{x}$ must be the global minimal solution $x^{*}$, $\tilde{x} = x^{*}$.
\label{convergence_theorem_2}
\end{theorem}  

If $1 - \lambda_k = 1/k$, for instance, 
	the series~(\ref{series_lambda}) is the harmonic series,
\[ 1 + \frac{1}{2} + \frac{1}{3} + \ldots + \frac{1}{k} + \ldots \]
The harmonic series is divergent.
Hence, with the choice of $\lambda_k = 1 - 1/k$, 
	if a consensus solution $\tilde{x}$ is found at some iteration 
	and it remains the same in the following iterations,
	it must be the global minimal solution $x^{*}$.
	
If $\{1 - \lambda_k\}_{k \ge 1}$, as another example, 
	is a convergent sequence of a positive limit,
	then $\sum_k 1 - \lambda_k$ is divergent.
In this case, a consensus solution is also the global minimal solution.
This statement can be generalized further to Cauchy sequences.
Every convergent sequence is a Cauchy sequence,
	and every Cauchy sequence is bounded.
Thus, if $\{1 - \lambda_k\}$ is a Cauchy sequence of a positive bound,
	a consensus solution is the global minimal solution.
	
To maximize the performance of a cooperative optimization algorithm,
	it is popular in the experiments to progressively increase the cooperation strength 
	as the iteration of the algorithm proceeds.
A weak cooperation level at the beginning leads to a fast convergence rate~(see Theorem~\ref{theorem_8}).
A strong cooperation level at a later stage of the iterations 
	increases the chance of finding a consensus solution.
Theorem~\ref{convergence_theorem_2} offers us some general guidance and justification 
	for choosing a variable cooperation strength.
It tells us that the increment of the cooperative strength
	should not be too fast if we want the guarantee of a consensus solution being the global optimal one.

\section{General Canonical Form of Cooperative Optimization}
\label{sec_gc_form}

By combining different forms of lower bound functions and different ways of decomposing objective functions,
	we can design cooperative optimization algorithms of different complexities and powers
	for attacking different optimization problems.
The basic canonical form of cooperative optimization~(\ref{cooperative_optimization_canonical}) 
	can be generalized further in a straightforward way to the general canonical one as follows.
	
Given a multivariate objective function $E(x_1, x_2, \ldots, x_n)$ of $n$ variables, 
	or simply denoted as $E(x)$, 
	where each variable is of a finite domain. 
Assume that $E(x)$ can be decomposed into $m$ sub-objective functions $E_i(x)$ which may satisfy the condition
\[ E(x) = \sum^{m}_{i=1} E_i(x) \ . \]

One may define another function $E_{-}(x)$, on the same set of variables as $E(x)$,
	as the composition of $m$ component functions as follows,
\[ E_{-}(x) = \sum^{m}_{i=1} \Psi_i (x^i) \ , \]
where $\Psi_i (x^i)$ is a component function defined on a subset of variables $X^{'}_i$, 
	$X^{'}_i \subset \{x_1, x_2, \ldots, x_n\}$, for $i=1,2,\ldots, m$.
$x^i$ is the restriction of $x$ on $X^{'}_i$, denoted as $x^i = x(X^{'}_i)$.

A cooperative optimization algorithm of the general canonical form
	is defined as minimizing the $m$ sub-objective functions $E_i(x)$ 
	in the following iterative and cooperative way,
\begin{equation}
\Psi^{(k)}_i (x^i) = \min_{X_i \setminus{X^{'}_i}}\left( E_i(x) + \lambda_k \sum^m_{j=1} w_{ij} \Psi^{(k-1)}_j(x^j)\right) \ , 
\label{cooperative_optimization_general}
\end{equation}
for $i=1,2,\ldots,m$.
In the equations, $k(=1,2,3,\ldots)$ is the iteration step; 
	$X_i$ is the set of variables contained in the functions at the right side of the $i$-th equation;
	$\lambda_k$ is a real value parameter at iteration $k$ satisfying $\lambda_k \ge 0$;
	and $w_{ij}~(1 \le i,j \le m)$ are also real value parameters satisfying $w_{ij} \ge 0$.
	
The solution at iteration $k$ is defined as
\[ \tilde{x}^{(k)} = \arg \min_{x} E^{(k)}_{-}(x) \ . \]
Moreover, $\tilde{x}^{(k)}$ is called a consensus solution 
	if it is the conditional optimum of all the $m$ minimization problems defined in (\ref{cooperative_optimization_general}).
That is, 
\[ \tilde{x}^{(k)}(X_i) = \arg \min_{x(X_i)} \left( E_i(x) + \lambda_k \sum^m_{j=1} w_{ij} \Psi^{(k-1)}_j(x^j)\right) \ , \]
when $x(X^{'}_i) = \tilde{x}^{(k)}(X^{'}_i)$ and $i=1,2,\ldots, m$.

One may choose the parameters $\lambda_k$ and $w_{ij}$ in such a way
	that they further satisfy the conditions of
	$\sum_i w_{ij} = 1$, for all $j$s,
	and all $\lambda_k$s are less than one ($\lambda_k < 1$).
With the settings, if the algorithm reaches its equilibrium at some iteration 
	and the solution of the iteration is also a consensus one,
	then it must be the global minimal solution
	(This global optimality condition can be proved in the exact same way as that of Theorem~\ref{sufficient_condition}).

The general canonical form can be further generalized to variable propagation matrices, 
	variable forms of lower bound functions,
	and variable ways of decomposing objective functions.
	
\section{Design Issues}	

A basic cooperative optimization algorithm~(\ref{cooperative_optimization}) (or a basic canonical one~(\ref{cooperative_optimization_canonical})) 
	is uniquely defined by the objective function decomposition $\{E_i(x)\}$, 
	the cooperation strength series $\{\lambda_k\}_{k \ge 1}$, 
	and the propagation matrix $(w_{ij})_{n \times n}$.
Some general guideline for designing the cooperation strength series 
	has discussed in the previous section.
This section focuses on the rest two.

\subsection{Objective Function Decomposition}

\subsubsection{Constraint Optimization Problems}

A large class of optimization problems have objective functions of the following form,
\begin{equation}
E(x_1,x_2,\ldots,x_n)=\sum^{n}_{i=1} f_i (x_{i}) + \sum_{(i,j) \in {\cal N}} f_{ij} (x_i, x_j) \ . 
\label{binary_cost_function}
\end{equation}
The function $f_i(x_i)$ is an unary function on variable $x_i$, for $i=1,2,\ldots, n$.
	and the function $f_{ij}(x_i, x_j)$ is a binary function on two variables $x_i, x_j$.
To note the collection of all defined binary functions,
	the set ${\cal N}$ is used which contains non-ordered pairs of variable indices
	where each pair $(i,j)$ corresponds to a defined binary function $f_{ij}(x_i, x_j)$.

The above optimization problems are also referred to as binary constraint optimization problems (binary COP) in AI.
The unary function $f_i(x)$ is called an unary constraint on variable $x_i$
	and the binary function $f_{ij}(x_i, x_j)$ is called a binary constraint on variables $x_i, x_j$.

Binary constraint optimization problems are a very general formulation 
	for many optimization problems arose from widely different fields.
Examples are the famous traveling salesman problems,
	weighted maximum satisfiability problems, 
	quadratic variable assignment problems, 
	stereo vision, 
	image segmentation, and many more.
Solving a binary constraint optimization problem is NP-hard in computation.

\subsubsection{Graphical Representation of Objective Functions}

An objective function in form of (\ref{binary_cost_function})
	can be represented with an undirected graph $G = (V,E)$.
In the graph, each variable $x_i$ is represented by a node, called a variable node,
	$V = \{x_1, \ldots, x_n\}$;
	each binary constraint $f_{i,j}(x_i, x_j)$ is represented by an undirected edge, 
	connecting the variable nodes $x_i$ and $x_j$, denoted by a non-ordered pair of variable nodes $(x_i, x_j)$.
By definition, the set $E$ of the edges of the graph $G$ is $E = \{(x_i, x_j)|(i,j) \in {\cal N} \}$.

\begin{figure}
\centerline{\epsfxsize 2.5cm \epsffile{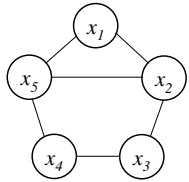}}
\caption{The graphical representation of the objective function of the simple example.}
\label{fig_constraint_graph}
\end{figure}

The simple example described in subsection~\ref{sec_simple_example}
	is a binary constraint optimization problem.
The objective function~(\ref{obj_simple_example}) of the simple example
 	has the form of (\ref{binary_cost_function}). 
It can be represented by an undirected graph as shown in Figure~\ref{fig_constraint_graph}. 

If edge $(x_i,x_j) \in E$, then the variable nodes $x_i, x_j$ are called neighbors to each other.
In graph theory, they are also called adjacent to each other.
Each variable node $x_i$ can have a number of neighboring variable nodes.
Let ${\cal N}(i)$ be the set of the indices of the neighboring variables of $x_i$. 
By definition, 
\[ {\cal N}(i) = \{j|(i,j) \in {\cal N} \} \ . \]

Using the notations, we can rewrite the objective function~(\ref{binary_cost_function}) as
\begin{equation}
E(x_1,x_2,\ldots,x_n)=\sum^{n}_{i=1} \left( f_i (x_{i}) + 1/2 \sum_{j \in {\cal N}(i)} f_{ij} (x_i, x_j) \right).
\label{binary_cost_function2}
\end{equation}

\subsubsection{Straightforward Decompositions}

The expression~(\ref{binary_cost_function2})
	for an objective function of a binary constraint optimization problem
	also defines a straightforward way to decompose the energy function.
That is,
\begin{equation}
E_i(x) =  f_i (x_{i}) + 1/2 \sum_{j \in {\cal N}(i)} f_{ij} (x_i, x_j), \quad \mbox{ for } i = 1, 2,\ldots, n \ .
\label{sub_binary_cost_function}
\end{equation}
Obviously, $\sum^{n}_{i=1} E_i(x) = E(x)$.
This kind of decompositions is so called the straightforward decompositions.
The sub-objective functions $E_i(x)$ in the straightforward decompositions can be easily minimized as 
\[ \min_{x} E_i(x) = \min_{x_i} \left(f_i (x_i) + 1/2 \sum_{j \in {\cal N}(i)} \min_{x_j} f_{ij} (x_i, x_j) \right) \ . \]

\begin{figure}
\centerline{\epsfxsize 8.0cm \epsffile{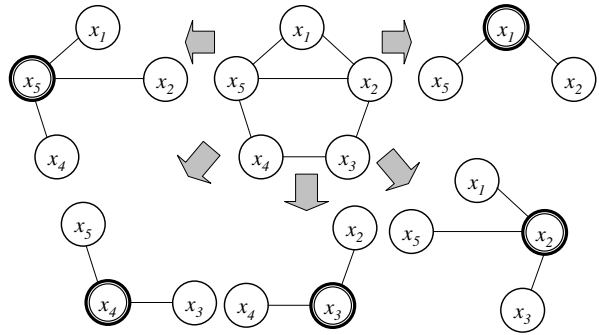}}
\caption{Decomposing a loopy graph into a number of loop-free subgraphs 
	based on the straightforward decompositions,
	one subgraph associated with each variable node (double circled).}
\label{fig_NeighborhoodDecomposition}
\end{figure}

Using the graphical representation of an objective function
	can help us to visualize the straightforward decomposition of an objective function.
For example, the decomposition of the objective function of the simple example presented in Subsection~\ref{sec_simple_example}
	can be viewed an instance of this kind of decompositions.
The original objective function has a graphical representation shown in Figure~\ref{fig_constraint_graph}.
Each sub-objective function $E_i(x)$ of the decomposition can also be represented by a graph,
	which must be a subgraph of the original one.
The graphical representation of the decomposition is illustrated in Figure~\ref{fig_NeighborhoodDecomposition}.
In the figure we can see that the original loopy graph is decomposed into 
	five loop-free subgraphs.

In general, in a graphical representation,
	the straightforward decompositions given in~(\ref{sub_binary_cost_function}) 
	can be understood as decomposing a loopy graph into $n$ loop-free subgraphs, 
	one subgraph is associated with each variable node in the original graph.
In the decomposition, 
	each subgraph is of a star-like structure with its associated variable node as the star center.
It consists of the variable node, 
	the neighbors of the node, and the edges connecting the node with its neighbors.
	
\subsubsection{Graph-Based Decompositions}

The graphical representation of an objective function may contain many loops.
That is the major cause of the difficulty at minimizing the objective function.
If the graph is loop-free, there exist algorithms with linear complexity (e.g., dynamic programming)
	that can minimize the objective function efficiently.
Therefore, if we can decompose an objective function with a loopy graphical representation
	into a number of sub-objective functions with loop-free graphical representations,
	a hard optimization problem is, thus, broken into a number of sub-problems of lower complexities.
Cooperative optimization can then be applied to solve those sub-problems in a cooperative way.
This kind of decompositions is called the graph-based decompositions.

It is important to note that 
	the modification given in (\ref{modified_objective_function}) for a sub-objective function
	does not change its graphical structure.
In other words, every modified sub-objective function defined in (\ref{modified_objective_function})
	has the exact same graphical structure as its original one.
This is because only unary functions, the assignment constraints $\Psi^{(k-1)}_i(x_i)$, 
	are introduced in the definition.
Therefore, any optimization algorithm applicable to the original sub-objective functions
	should also be applicable to the modified ones.
In other words,
	if a sub-objective function is of a tree-like structure,
	then its modified version defined by (\ref{modified_objective_function})
	must have the exact same tree-like structure.
Both of them can be minimized efficiently via dynamic programming.

\subsubsection{Spanning-Tree Based Decompositions}

In terms of the graph based decompositions,
	the straightforward decompositions are based on the direct neighbors of each variable node.
Another possible way of decomposing an objective function
	is based on the spanning trees of the graph representing the function.
A tree is called a spanning tree of an undirected graph $G$ if it is a subgraph of $G$
	and containing all vertices of $G$.
Every finite connected graph $G$ contains at least one spanning tree $T$.

\begin{figure}
\centering
\includegraphics[width=8.0cm]{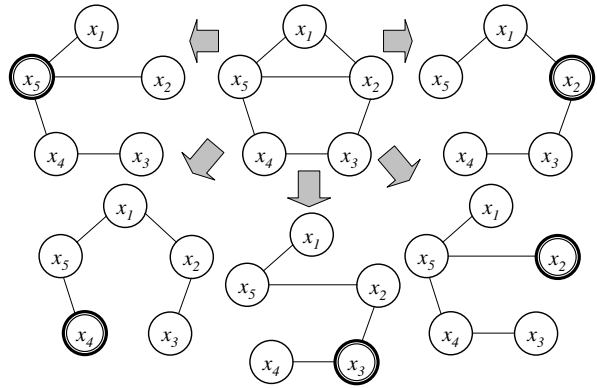}
\caption{Decomposing a loopy graph into a number of spanning trees,
	one spanning tree associated with each variable node (double circled).}
\label{fig_SpanningTreeDecomposition}
\end{figure}

Given an objective function $E(x)$ of $n$ variables in form of (\ref{binary_cost_function}).
Let $G = (V, E)$ be its graphical representation with $n$ variable nodes.
Without loss of generality, we assume that $G$ is connected 
	(otherwise it implies that the original minimization problem can be broken into several independent sub-problems).
For each variable node $x_i$ of $G$, 	
	we can associate a spanning tree of $G$, denoted as $T_i = (V, E_i)$ ($T_i$ shares the same set of nodes as $G$).
There are $n$ such spanning trees in total, $T_1, T_2, \ldots, T_n$ (some trees may possibly be duplicated).
We also choose those $n$ spanning trees in a way such that each edge of $G$
	is covered at least by one of the $n$ trees.
Figure~\ref{fig_SpanningTreeDecomposition} shows an example
	of decomposing the graph representing the objective function of the simple example
	into five spanning trees.

After decomposing the graph $G$ into $n$ spanning trees,
	if we can define a sub-objective function $E_i(x)$ for each spanning tree such that
\begin{enumerate}
\item $E(x) = \sum_i E_i(x)$;
\item The graphical representation of $E_i(x)$ is $T_i$, for $i=1,2,\ldots,n$.
\end{enumerate}
Then the set $\{E_i(x)\}$ is a legitimate decomposition of the original objective function $E(x)$.
This kind of decompositions is so called the spanning-tree based decompositions.

Given an objective function $E(x)$ of a binary constraint optimization problem
	with a graphical representation $G$ and $n$ spanning trees,
	each unary constraint $f_j(x_j)$ of $E(x)$ is associated with a variable node $x_j$ in $G$
	covered by all the $n$ spanning trees.
Each binary constraint $f_{jk}(x_j,x_k)$ of $E(x)$ is associated with an edge $(x_j, x_k)$ in $G$
	covered at least by one of the $n$ spanning trees.
Assume that the edge $(x_j, x_k) \in E$ is covered by $m(j,k)$ spanning trees, 
	where $m(j,k)$ should be a positive integer.
One way for defining sub-objective functions $E_i(x)$ to satisfy the above two conditions is 
	given as follows,
\begin{equation}
E_i(x) = \frac{1}{n} \sum^n_{j=1} f_j (x_j) + \sum_{(j,k) \in E_i} \frac{1}{m(j,k)} f_{jk}(x_j, x_k) \ , 
\label{tree_decomposition}
\end{equation}
for $i=1,2,\ldots, n$.

\subsubsection{Some Properties of Spanning-Tree Based Decomposition}

We can apply algebraic graph theory to reveal some of properties of graphs and their spanning trees.
Let $G = (V,E)$ be a finite simple graph of $n$ nodes, with vertices $v_1, \ldots, v_n$.
If $G$ is a graphical representation of an objective function, 
	then the vertices $v_i$ is the variable node $x_i$, i.e., $v_i = x_i$.

The connectivity of a graph $G$ can be represented by the adjacency matrix $A$ of $G$.
The adjacency matrix $A$ of $G$ is defined as
\[ A = (a_{ij})_{n \times n} = \left\{ \begin{array} {ll}
                               1, & \mbox{if  $(v_i, v_j) \in E$}, \\ 
                               0, & \mbox{if $(v_i, v_j) \not \in E$}. 
                               \end{array}
                       \right. \]       
The adjacency matrix of an undirected graph is symmetric.

Let $Q$ be the adjacency matrix of $G$ where the diagonal entries $Q_{ii}$ are replaced by the degrees of vertices $-deg(v_i)$.
Let $Q^{*}$ be the matrix obtained by removing the first row and column of $Q$. 
Then the number of spanning trees in G is equal to $|det(Q^{*})|$ (Kirchoff's Matrix-Tree Theorem).
Particularly, if $G$ is a complete graph, 
	$Q^*$ has the determinant $n^{n-2}$.
That is, every complete graph with $n$ vertices ($n > 1$) has exactly $n^{n-2}$ spanning trees (Theorem of Cayley).

\subsubsection{Further Generalization of Graph-Based Decompositions}

Using factor graph~\cite{Kschischang01}, 
	any objective function containing $k$-ary constraints, where $k$ can be any integer number,
	can be represented as a graph.
With the representation, the two aforementioned kinds of decompositions can be easily generalized further
	to decompose the objective function.
Special decompositions can also be explored for graphs with special structures 
	or constraints of some special properties
	to maximize the power of cooperative optimization algorithms.
Another way to apply the two kinds of decompositions for the $k$-ary constraints case
	is by converting the constraints of orders higher than two into 
	binary constraints via variable clustering technique.

\subsection{The Propagation Matrix}

As mentioned in the previous subsection,
	the objective function of a binary constraint optimization problem
	has a graphical representation $G = (V, E)$.
For the straightforward decompositions described in the previous subsection,
	we can design a propagation matrix
	based on the adjacency of the graph $G$ as follows  
\begin{equation}
W = (w_{ij})_{n \times n} = \left\{ \begin{array} {ll}
                               1/d_j, \quad & \mbox{if  $(x_i, x_j) \in E$}, \\ 
                               0, \quad & \mbox{otherwise}. 
                               \end{array}
                       \right. 
\label{p1}
\end{equation}
where $d_j$ is the degree of the variable node $x_j$.
The propagation matrix~(\ref{p2}) of the simple example 
	is designed in this way.

Another way to design a propagation matrix is given as follows,  
\begin{equation}
W = (w_{ij})_{n \times n} = \left\{ \begin{array} {ll}
                               1/(d_j+1), \quad & \mbox{if  $(x_i, x_j) \in E$ or $i = j$}, \\ 
                               0, \quad & \mbox{otherwise}. 
                               \end{array}
                       \right.  
\label{p3}
\end{equation}
Such a matrix has all of the diagonal elements of non-zero values.

\subsection{Cooperative Optimization in Simple Form}	

The design of cooperative optimization algorithms
	is not trivial even with the aforementioned guidelines.
In the basic canonical form~(\ref{cooperative_optimization_canonical}),
	there are $n \times n$ values for the propagation matrix $(w_{ij})_{n \times n}$ 
	and a series of values for the cooperation strength $\lambda_k$. 
To ease the design job for engineers and practitioners,
	the difference equations~(\ref{cooperative_optimization_canonical}) of the basic canonical form
	can be simplified to 
\begin{equation}
\Psi^{(k)}_i (x_i) = \min_{X_i \setminus{x_i}}\left( E_i(x) + \alpha \sum_{j} \Psi^{(k-1)}_j(x_j) \right),
	\quad \mbox{for $i=1,2,\ldots,n$} \ , 
\label{simple_cooperative_optimization0}
\end{equation}
where $\alpha$ is the only parameter to be tuned in experiments to maximize performance.
It plays the same role as the cooperation strength $\lambda_k$
	for controlling the cooperation level among the agents in a system. 
The above set of simplified difference equations defines the simple form of cooperative optimization.

The simple form is derived from the basic canonical form~(\ref{cooperative_optimization_canonical})
	by setting $w_{ij}$ be a positive constant $w$, 
	for any $i$ and $j$, if $x_j$ is contained in $E_i(x)$; and $w_{ij} = 0$, otherwise.
We also let the cooperation strength $\lambda_k$ be of a constant value $\lambda$. 
Let $\alpha = \lambda w $, we have (\ref{cooperative_optimization_canonical}) simplified to (\ref{simple_cooperative_optimization0}).

If the parameter $\alpha$ is of a large value,
	the difference equations~(\ref{simple_cooperative_optimization0}) of a simple cooperative optimization algorithm
	may have value overflow problems in computing the assignment constraints $\Psi^{(k)}_i (x_i)$.
To improve its convergence property,
	we can offset each $\Psi^{(k)}_i (x_i)$ by a value at each iteration.
One choice is the minimal value of $\Psi^{(k)}_i(x_i)$. 
That is we offset $\Psi^{(k)}_i(x_i)$ by its minimal value as follows, 
\[\Psi^{(k)}_i(x_i) := \Psi^{(k)}_i(x_i) - \min_{x_i} \Psi^{(k)}_i(x_i) \ . \]
Thus, the offsetting defines an operator on $\Psi^{(k)}_i(x_i)$, denoted as ${\cal O} (\Psi^{(k)}_i(x_i))$.
With the notation, the difference equations of a simple cooperative optimization algorithm become
\begin{equation}
\Psi^{(k)}_i (x_i) = {\cal O}\left(\min_{X_i \setminus{x_i}}\left( E_i(x) + \alpha \sum_{j} \Psi^{(k-1)}_j(x_j) \right)\right),
	\quad \mbox{for $i=1,2,\ldots,n$} \ . 
\label{simple_cooperative_optimization}
\end{equation}

\section{A Case Study in Computer Vision}
Just like many other problems in computer vision and image processing, 
	stereo matching can be formulated as a binary constraint optimization problem
	with an objective function $E(x)$ in form of (\ref{binary_cost_function}).
For detail about the energy function definitions 
   used for stereo matching, please see~\cite{Scharstein2002}.   
Basically,
   an unary constraint $f_i(x_i)$ in~(\ref{binary_cost_function}) measures 
   the difference of the intensities between site $i$ from one image and its corresponding site in another image
   given the depth of the site.
A binary constraint $f_{ij} (x_i, x_j)$ measures the difference 
   of the depths between site $i$ and site $j$.
This type of constraints is also referred to as the smoothness constraint in literature.
It has also been widely used in solving image segmentation and other vision tasks.

In our experiments, we apply the simplified form of cooperative optimization~(\ref{simple_cooperative_optimization})
	for stereo matching with the parameter $\alpha$ is set to $0.16$.
The maximum number of iterations is set to $16$.
The objective function associated with stereo-matching 
	is decomposed based on the spanning-tree based decomposition.
The detail of the decomposition 
	and the minimization of the sub-objective functions are offered in the following subsection.
	
\subsection{Decomposing Grid-like Graphs}

Often times, the graphical representation of the objective function of
	an image segmentation problem or a stereo matching problem is of a 2-D grid-like structure.
Because a 2-D grid-like graph is highly regular in structure,
	its spanning trees can be easily defined in a systematic way.

Given an objective function of a 2-D grid-like graphical representation $G = (V,E)$, 
	let $M$ be the height of the grid (the number of rows)
	and $N$ be the width of the grid (the number of columns).
Let $E^h$ be the set of all horizontal edges of $G$ and $E^v$ be the set of all vertical edges.
There are in total $M \times N$ nodes, one for each variable.
There are in total $M (N-1)$ horizontal edges and $N (M-1)$ vertical edges.
With the notations, the objective function can be expressed as
\[ E(x) = \sum_{i \in V} f_i(x_i) + \sum_{(i,j) \in E} f_{ij}(x_i, x_j) \ , \]
or equivalently,
\[ E(x) = \sum_{i \in V} f_i(x_i) + \sum_{(i,j) \in E^h} f_{ij}(x_i, x_j) + \sum_{(i,j) \in E^v} f_{ij}(x_i, x_j) \ . \]
	
The horizontal path $P^h_i = (V^h_i, E^h_i)$ through a variable node $x_i$ consists of all the nodes
	at the same horizontal line as $x_{i}$,
	together with the edges connecting those nodes. 
The vertical path $P^h_i = (V^v_i, E^v_i)$ through a variable node $x_i$ consists of all the nodes
	at the same vertical line as $x_{i}$, 
	together with the edges connecting those nodes. 

\begin{figure}
\begin{center}
\includegraphics[width=3.9cm]{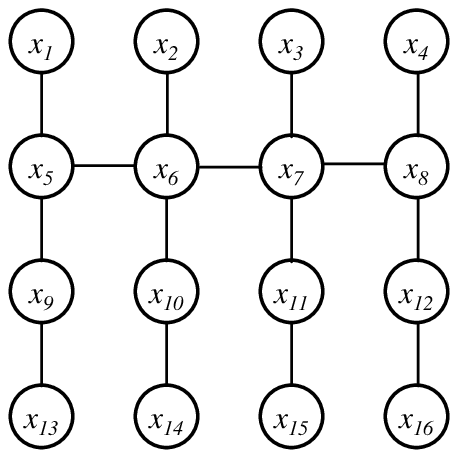}
\includegraphics[width=3.9cm]{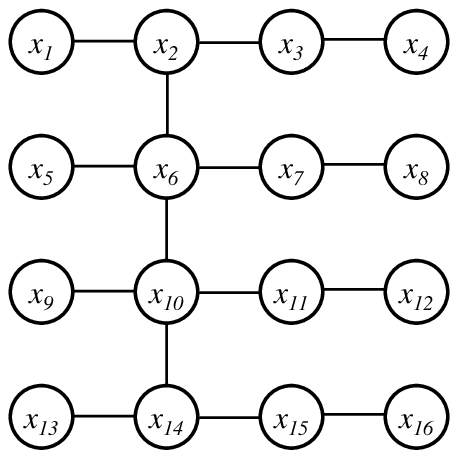}
\end{center}
\caption{A horizontal spanning tree (left) and a vertical spanning tree (right) 
	with the variable node $x_{6}$ as their roots.}
\label{fig_GridGraphHSpan}
\end{figure}

For each variable node $x_i$, let us define two spanning trees with the node as the root,
	called the horizontal spanning tree $T^h_i$ and the vertical spanning tree $T^v_i$, respectively.
The horizontal spanning tree $T^h_i$ consists of the horizontal path through the variable node $x_i$
	and all the vertical paths through each node in the horizontal path.
The vertical spanning tree $T^v_i$ consists of the vertical path through the variable node $x_i$
	and all the horizontal paths through each node in the vertical path
 	(the illustrations are shown in Fig.~\ref{fig_GridGraphHSpan}).

Let the functions $E^h_i(x),E^v_i(x)$ be the objective functions 
	associated with the horizontal spanning tree $T^h_i$ and the vertical spanning tree $T^v_i$, respectively.
Following the general design guideline described in the previous section for the spanning-tree based decompositions
	(see Eq.~(\ref{tree_decomposition})),
	we can define $E^h_i(x)$ and $E^v_i(x)$ as
\[E^h_i(x) = a \sum_{i^{'} \in V} f_{i^{'}} (x_{i^{'}}) + 
	b \sum_{(i^{'},j) \in E^h_i} f_{i^{'}j} (x_{i^{'}}, x_j) + c \sum_{(i^{'},j) \in E^v} f_{i^{'}j} (x_{i^{'}}, x_{j}) \ , 
\]
\[
E^v_i(x) = a \sum_{i^{'} \in V} f_{i^{'}} (x_{i^{'}}) + 
	c \sum_{(i^{'},j) \in E^v_i} f_{i^{'}j} (x_{i^{'}}, x_j) + b \sum_{(i^{'},j) \in E^h} f_{i^{'}j} (x_{i^{'}}, x_{j})\ , 
\]
where $a = 1/2MN$, $b = 1/(MN+N)$, and $c=1/(MN+M)$.

The sub-objective function $E_i(x)$ associated with variable $x_i$ is defined as
\begin{equation}
E_i(x) = E^h_i(x) + E^v_i(x) \ . 
\label{hv_decomposition}
\end{equation}
Clearly, we have $\sum_i E_i(x) = E(x)$.

\begin{figure}
\begin{center}
\includegraphics[width=9.5cm]{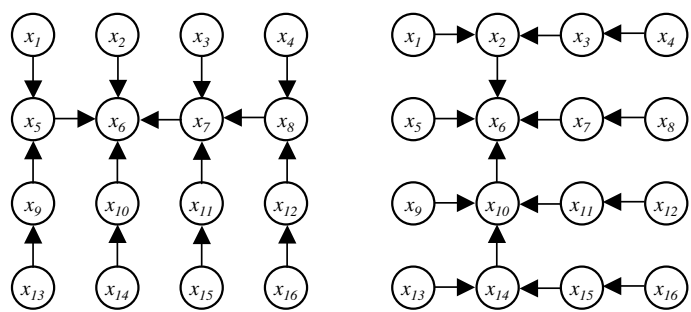}
\caption{The message flows among the variable nodes when applying the dynamic programming technique 
	for minimizing the objective functions associated with the horizontal spanning tree (left) and the vertical spanning tree (right)
	with the variable node $x_{6}$ at their roots.}
\label{fig_GridGraphMessageFlow}
\end{center}
\end{figure}

As mentioned before, any objective function of a tree-like structure
	can be minimized efficiently using the dynamic programming technique.
It is of a linear computational complexity 
	and is simply based on local message passing from the leave nodes all the way back to the root node.
The books~\cite{MacKay:itp,Pearl88} offer a detail explanation 
	about message passing and the dynamic programming technique.
When applying the technique for minimizing the objective function of a horizontal or vertical spanning tree,
	the message flows among the variable nodes are illustrated in Figure~\ref{fig_GridGraphMessageFlow}.

\subsection{Experimental Results}

\begin{figure}
\center{{\epsfxsize 3.3cm \epsffile{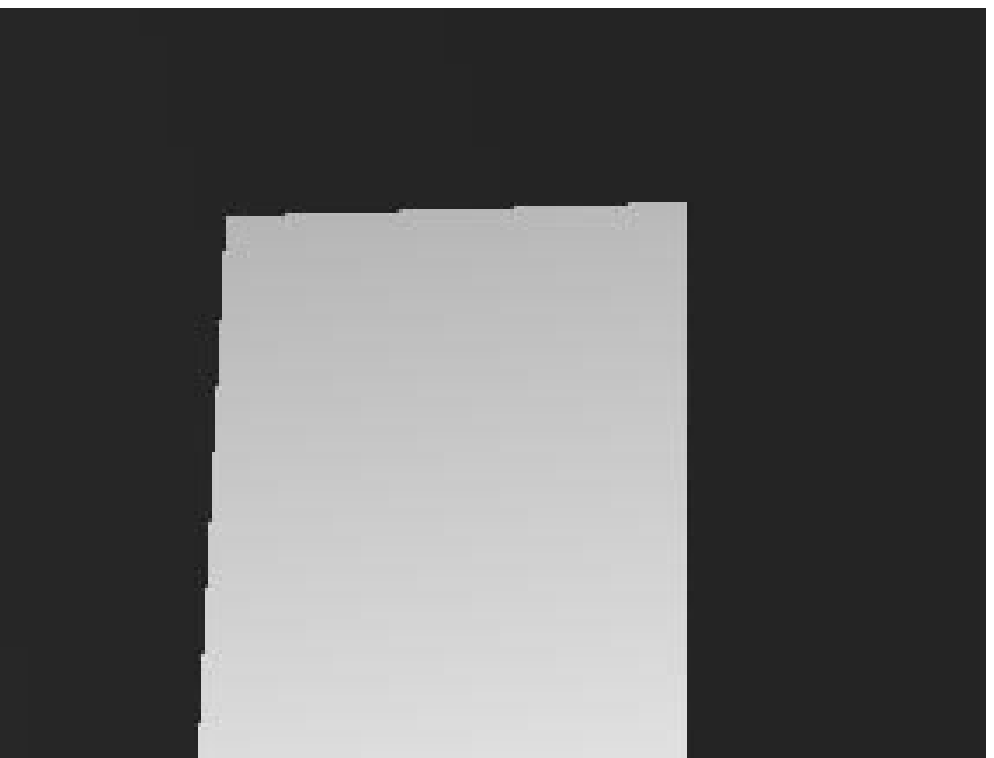}} {\epsfxsize 3.3cm \epsffile{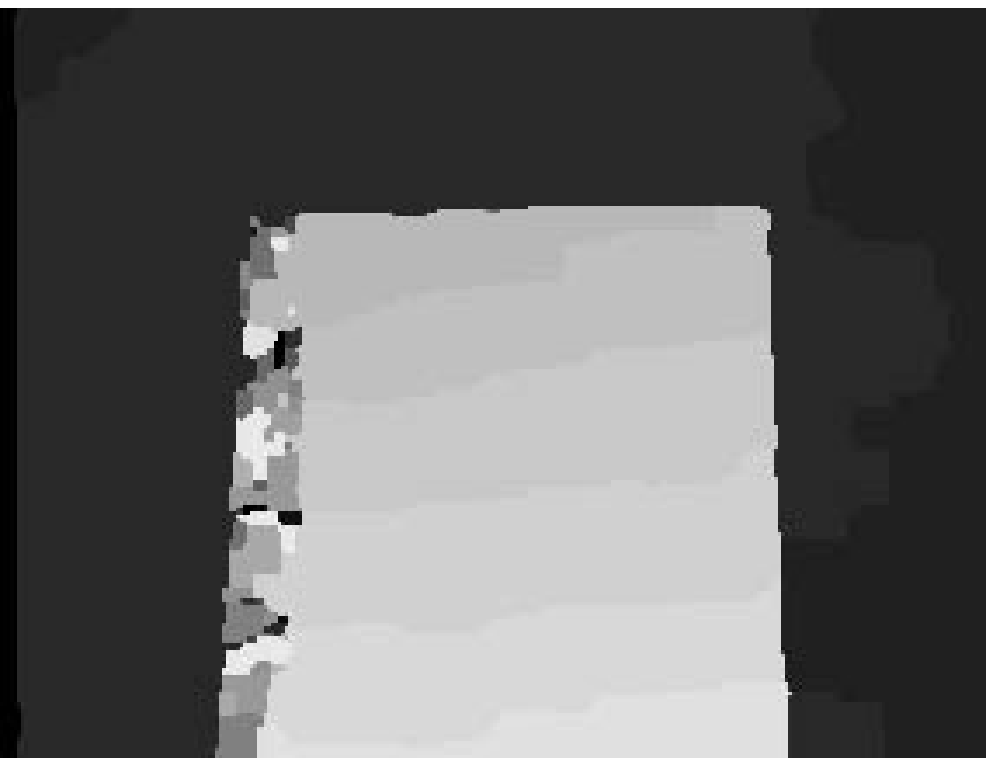}} {\epsfxsize 3.3cm \epsffile{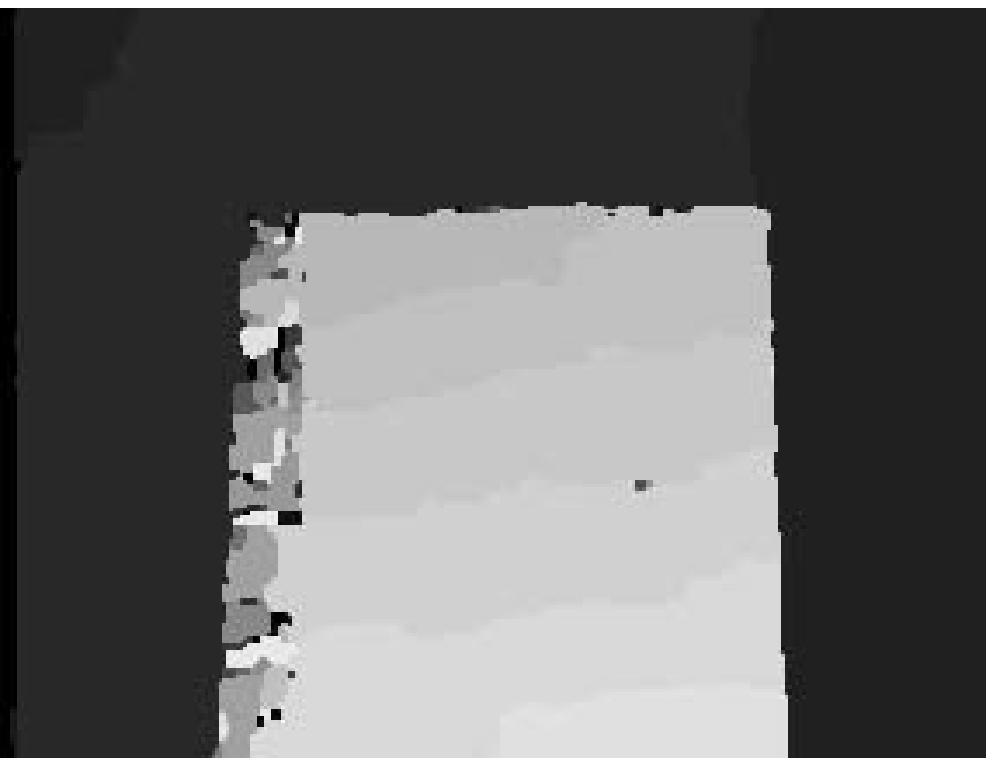}}}
\center{{\epsfxsize 3.3cm \epsffile{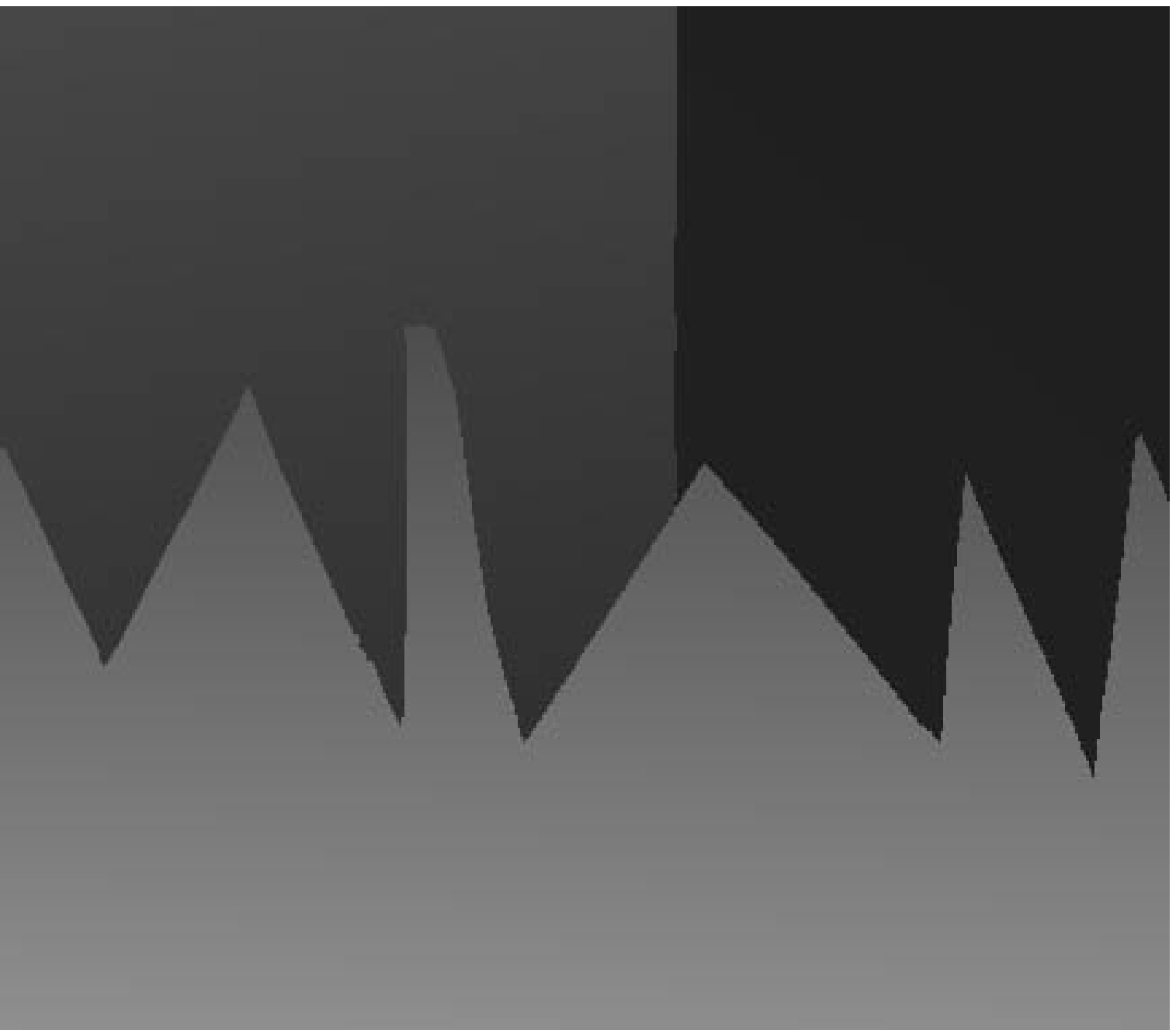}} {\epsfxsize 3.3cm \epsffile{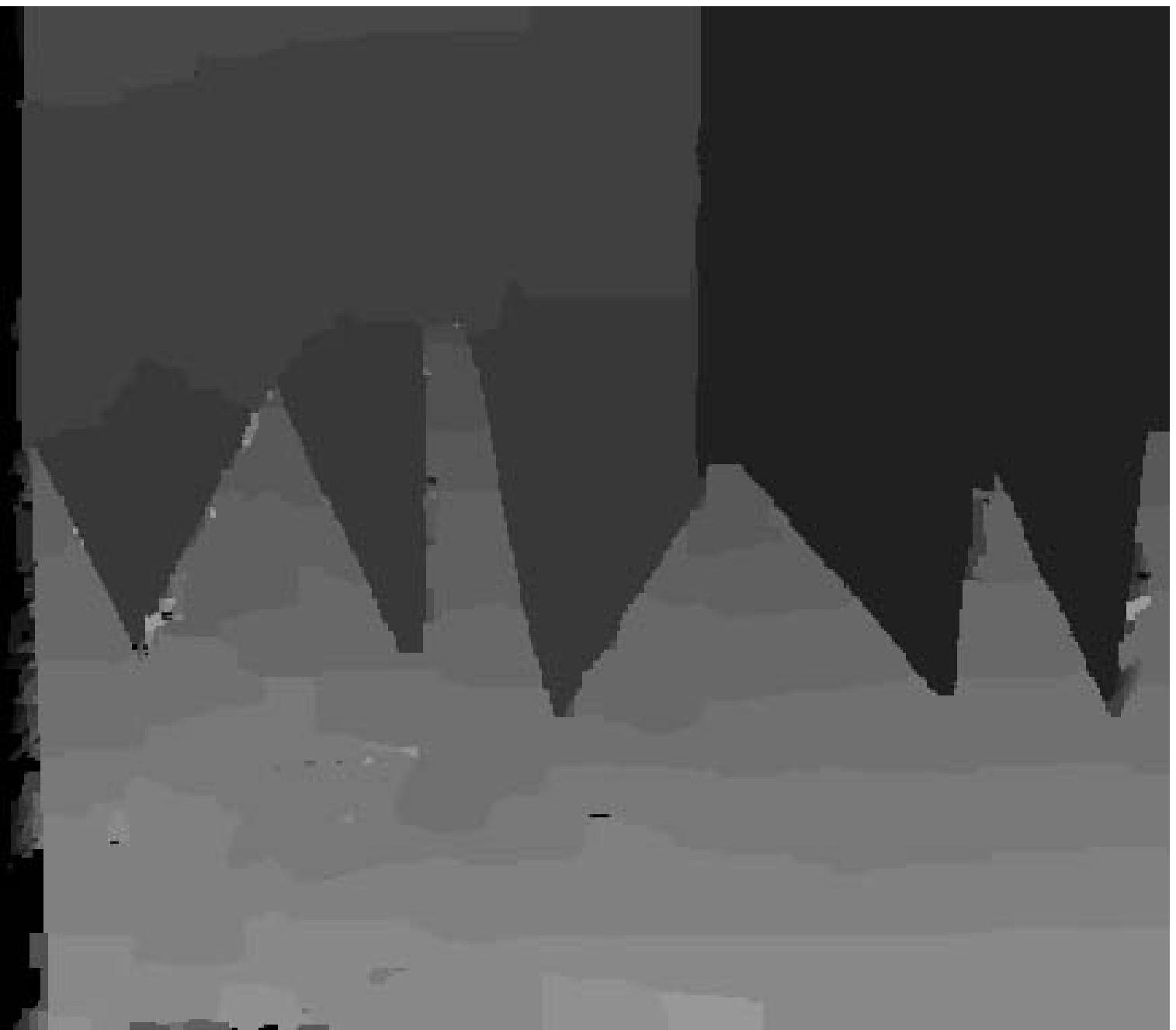}} {\epsfxsize 3.3cm \epsffile{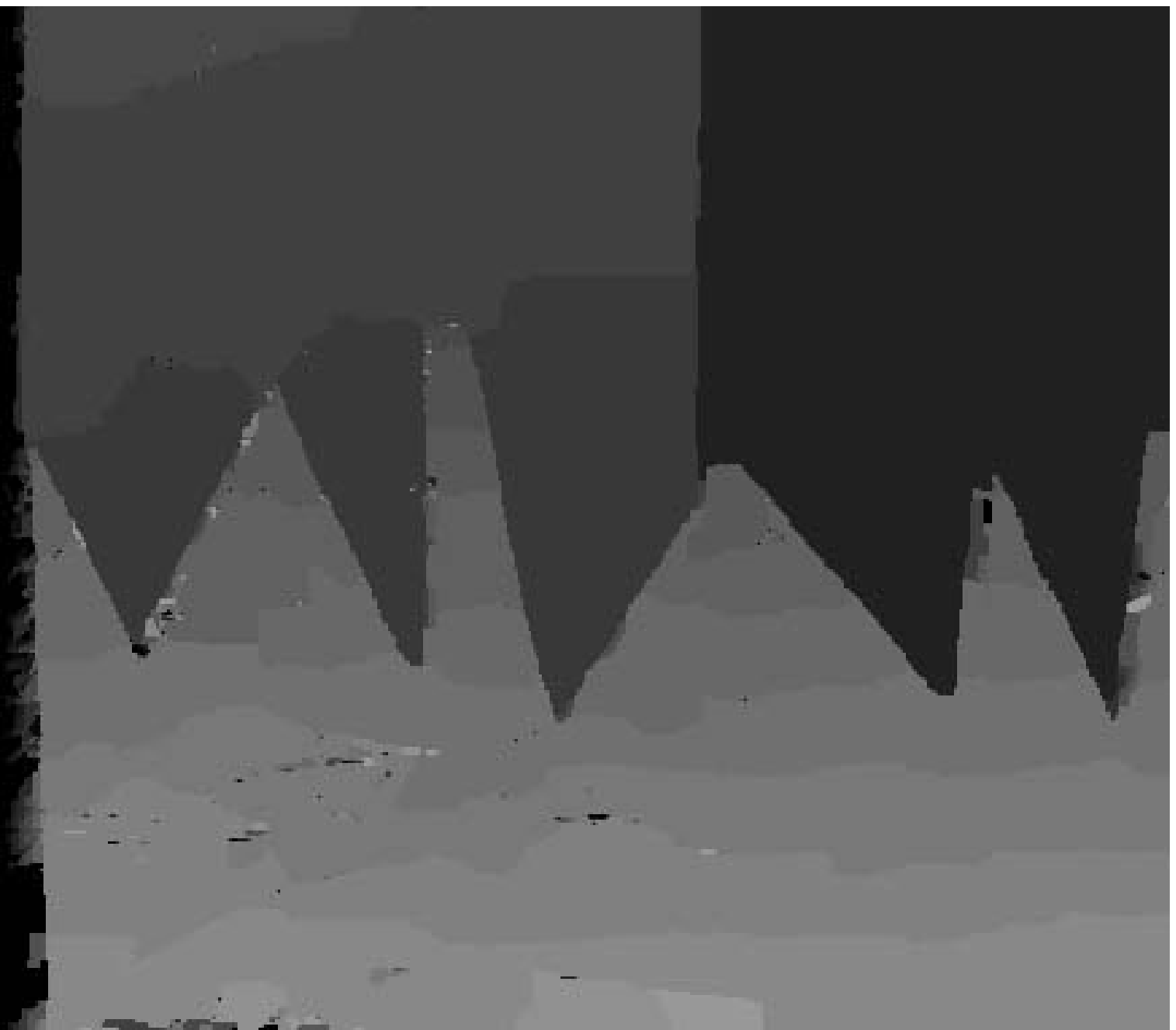}}}
\center{{\epsfxsize 3.3cm \epsffile{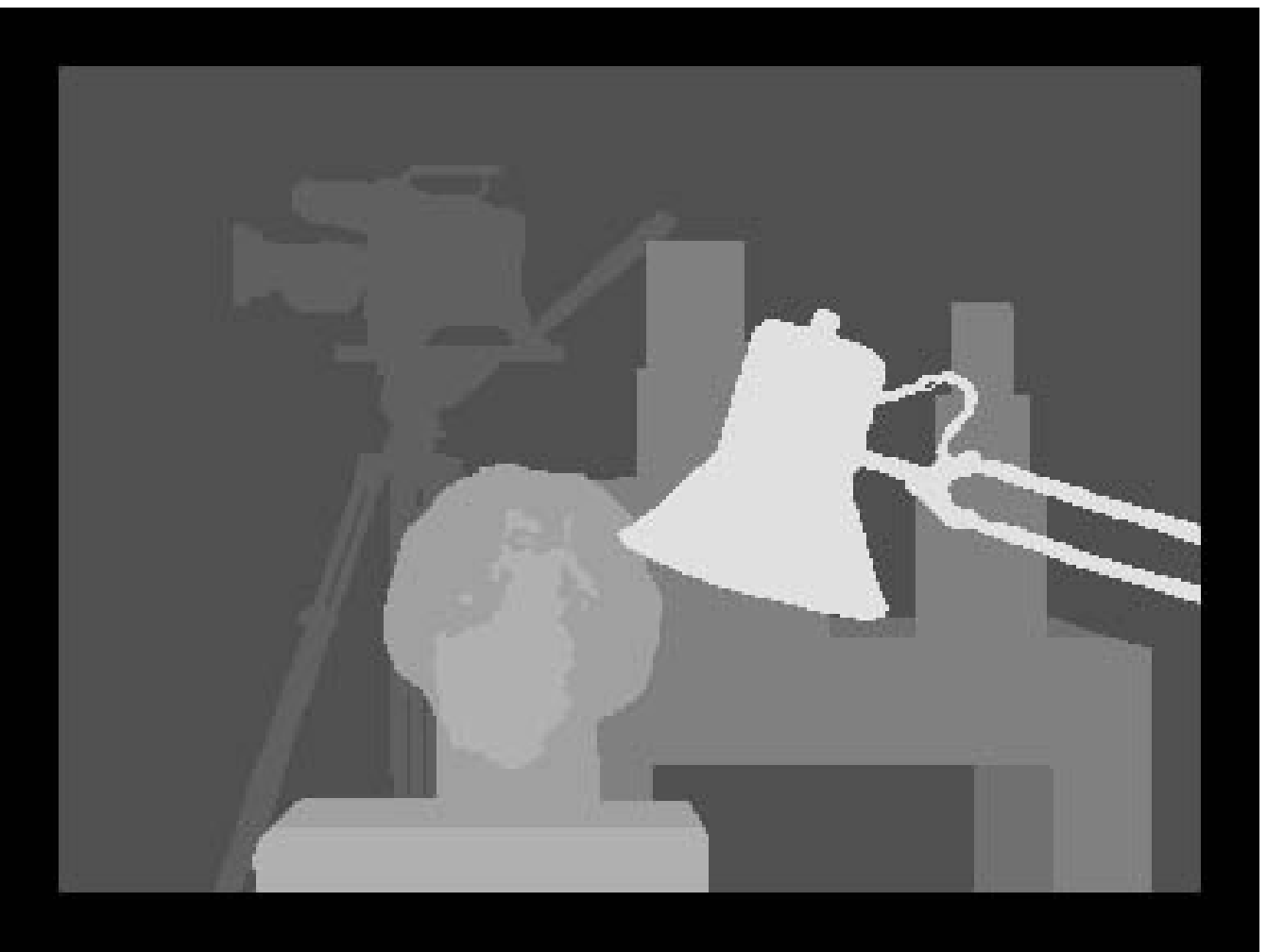}} {\epsfxsize 3.3cm \epsffile{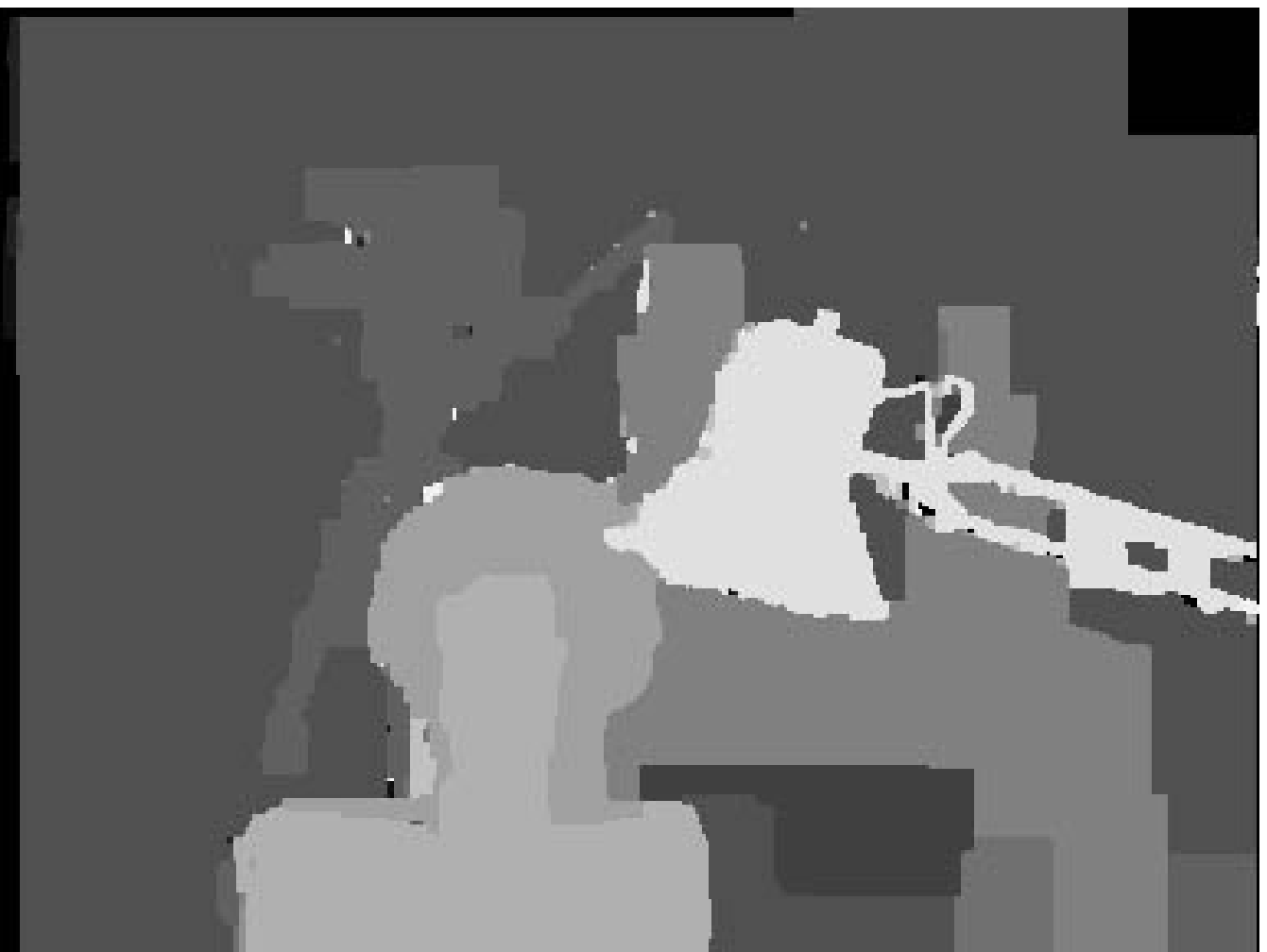}} {\epsfxsize 3.3cm \epsffile{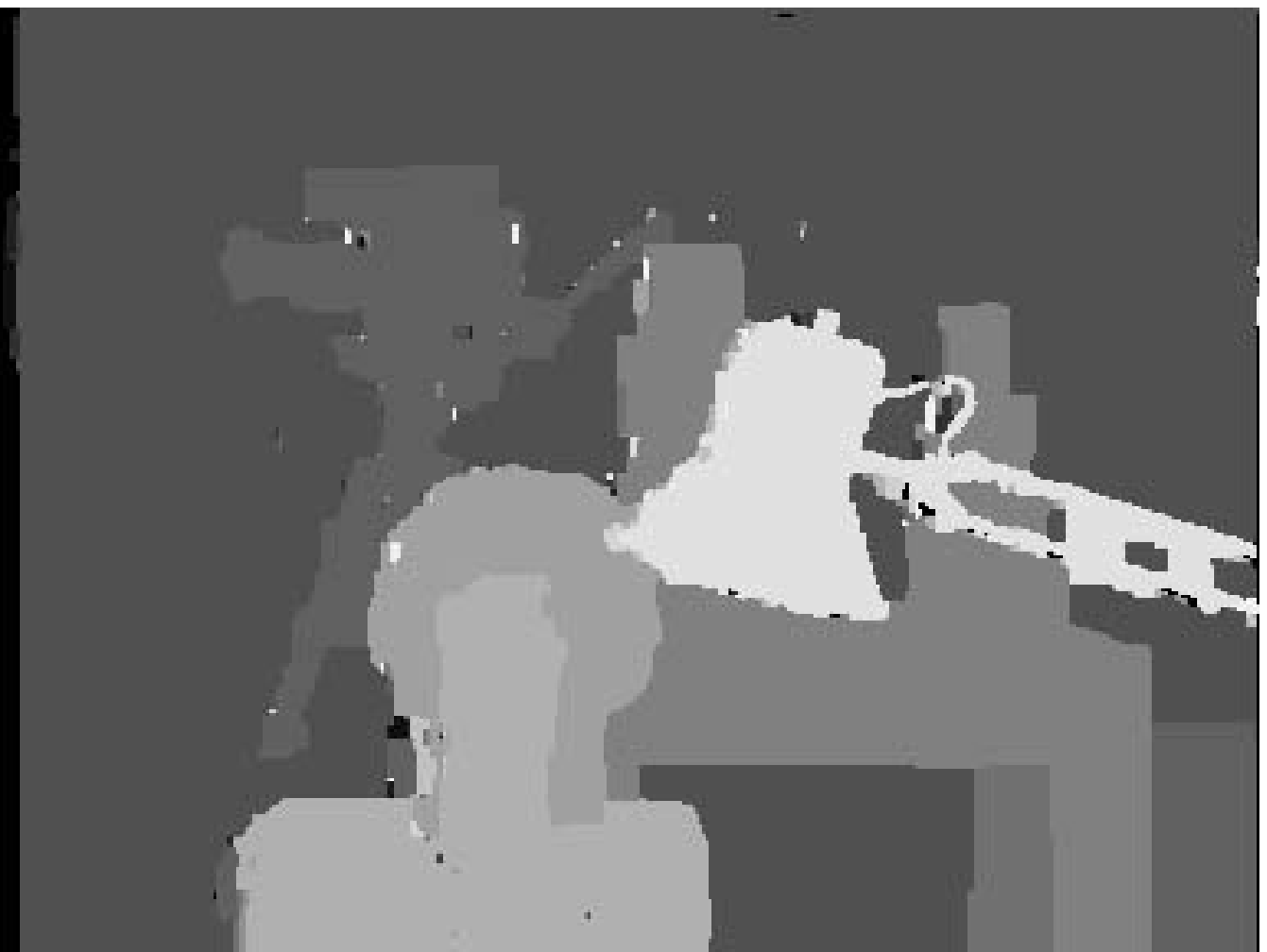}}}
\center{{\epsfxsize 3.3cm \epsffile{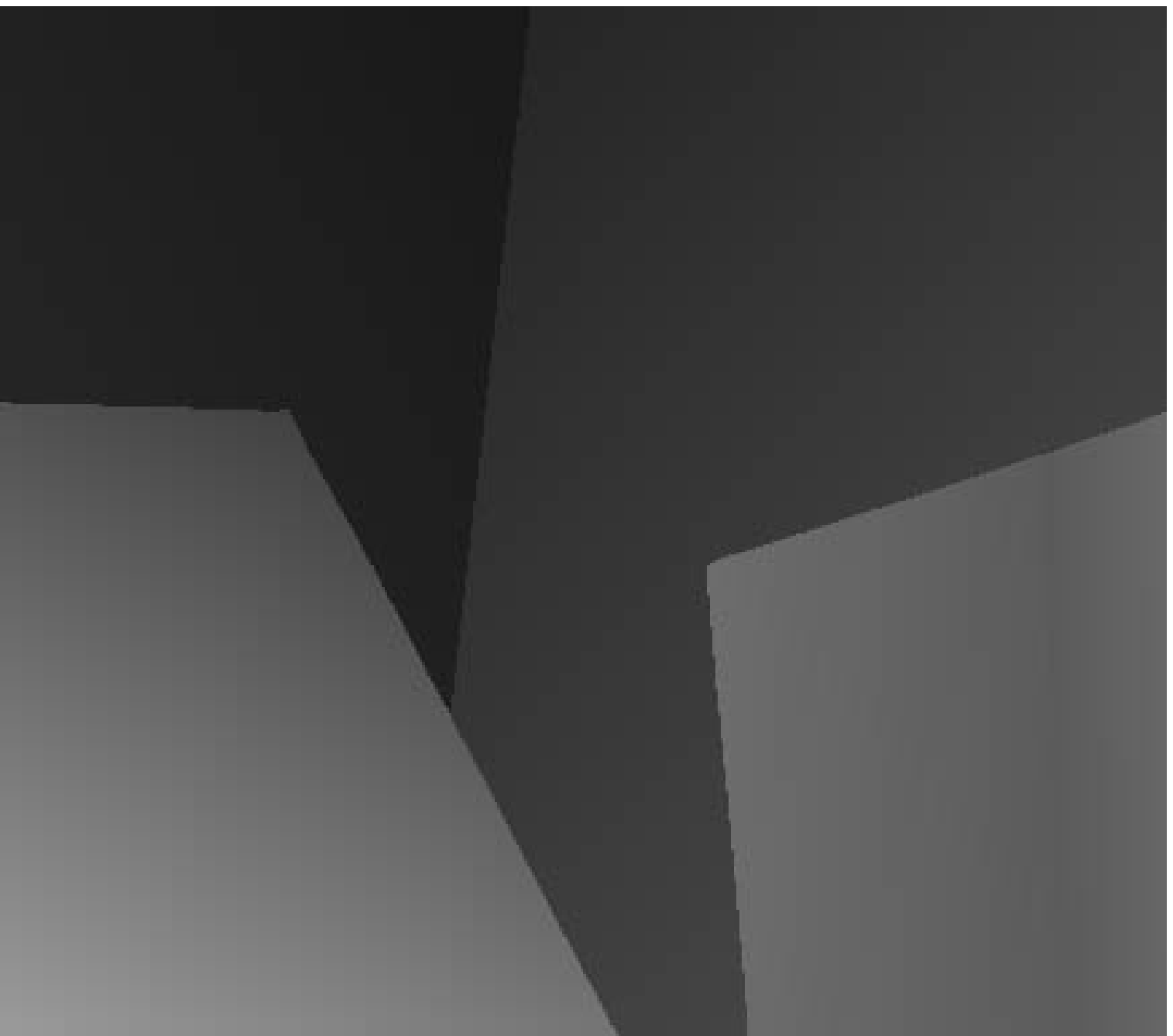}} {\epsfxsize 3.3cm \epsffile{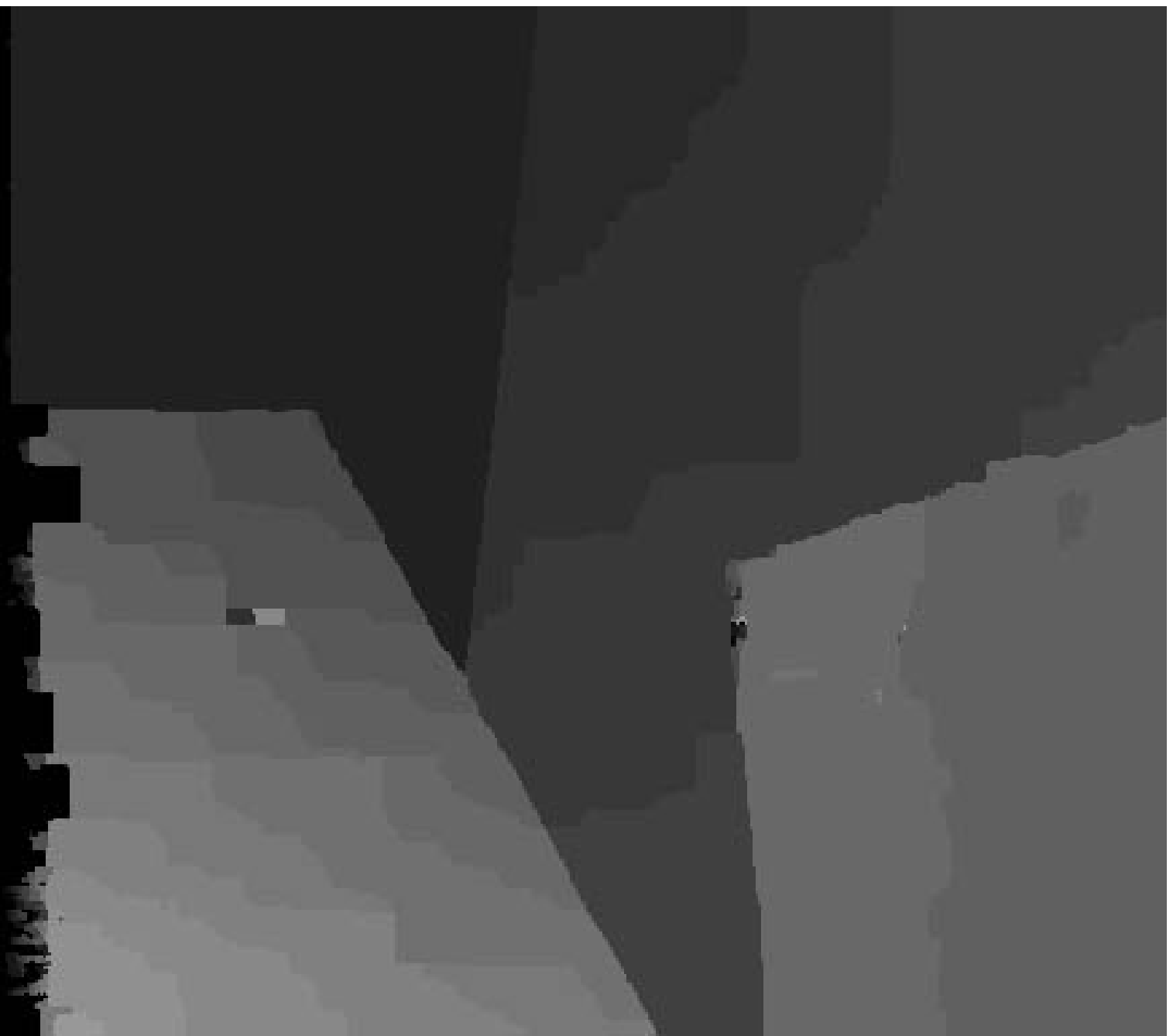}} {\epsfxsize 3.3cm \epsffile{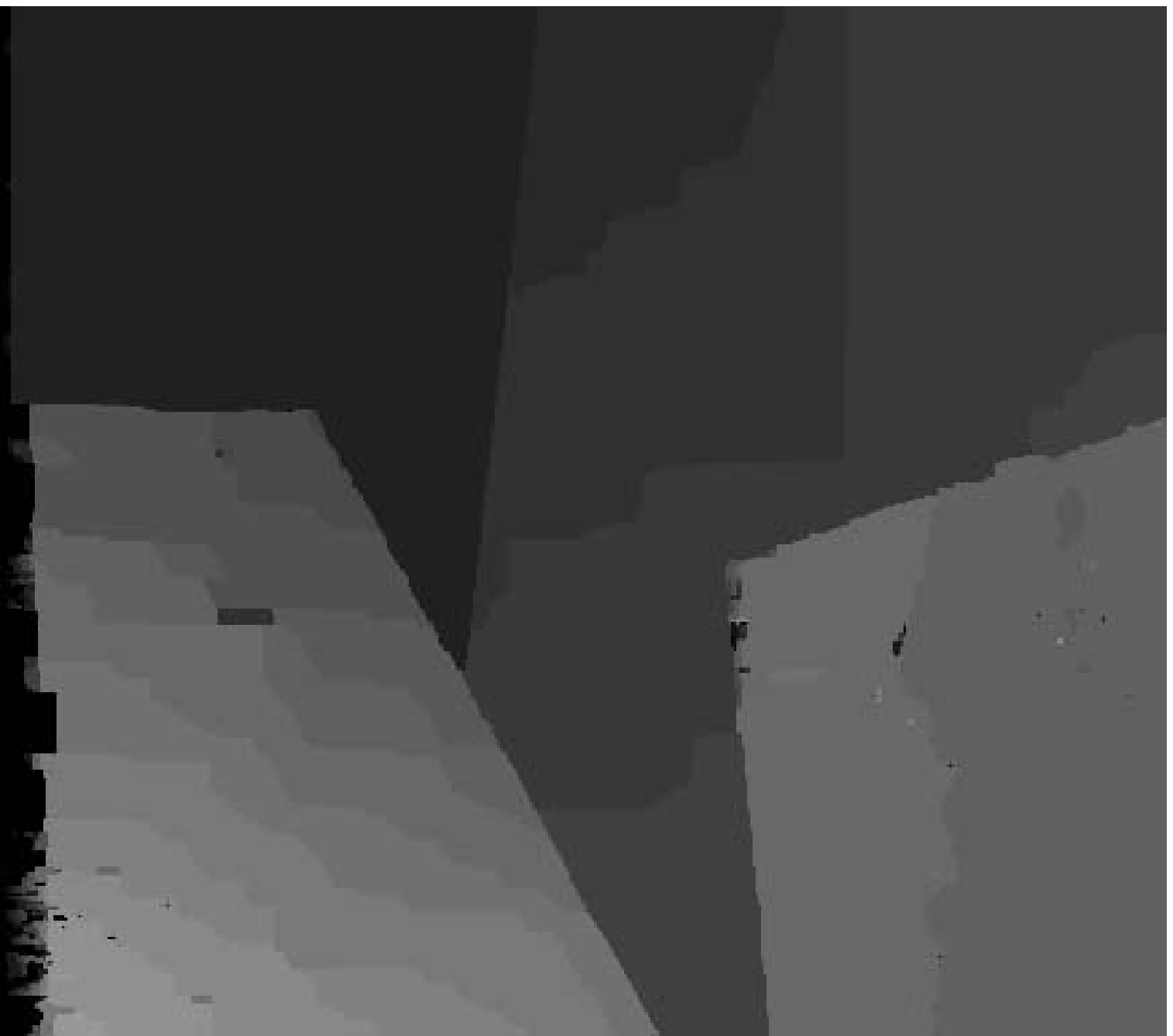}}}
\caption{The ground truths (the left column), cooperative optimization (the middle column), and  graph cuts (the right column).}
\label{fig_cc}
\end{figure}

The Middlebury College evaluation framework~\cite{Scharstein2002} for stereo matching is used in the experiments.
The script used for evaluation is based on {\it exp6\_gc.txt} offered in the framework.
The other settings come from the default values in the framework.
The results of stereo matching algorithms together with the ground truths for the four test stereo image pairs 
	from the evaluation framework are shown in Figure~\ref{fig_cc}. 
The quality of solutions of both algorithms are very close to each other from a visual inspection.

The following four tables show the performance of the cooperative optimization algorithm (upper rows in a table)
   and the graph cut algorithm (lower rows in a table) over the four test image sets.
The solution quality is measured in the overall area, no occluded areas, occluded areas, 
	textured areas, texture-less areas, and near discontinuity areas
	(see \cite{Scharstein2002} for the detail description of the evaluation framework).
Both algorithms does not handle occluded areas (an occluded area is one that is visible in one image, but not the other).
Also, the runtimes of the two algorithms (co = cooperative optimization algorithm, gc = graph cuts) are listed.
From the tables we can see that the cooperative optimization is very close to graph cuts in terms of solution quality and energy states.
However, the cooperative optimization algorithm is around $20$ times faster than graph cuts in the software simulation.

\begin{center}
{\small
\begin{tabular}{ccccccc}
\hline
\multicolumn{7}{c}{image = Map} \\
\multicolumn{3}{c}{time: co=17s / gc=337s} & \multicolumn{4}{c}{energy: co=328,658 / gc=321,144} \\
\hline
& {\tiny ALL} &  {\tiny NON OCCL} &  {\tiny OCCL} &  {\tiny TEXTRD} & {\tiny TEXTRLS} & {\tiny D\_DISCNT }\\
Error & 4.04 &   0.85 &  16.18  &  0.85 &  0.36   &  2.84  \\
Bad Pixels & 5.35\% &  0.18\% & 86.78\% &  0.18\% &  0.00\% &  2.46\% \\
\hline
Error &    3.91  &  1.07 &  15.45 &   1.07  &  0.38  &  3.65 \\
Bad Pixels &    5.63\% &  0.36\% & 88.76\% &  0.36\% &  0.00\% &  4.52\% \\
\hline
\end{tabular}

\vskip 0.3in

\begin{tabular}{ccccccc}
\hline
\multicolumn{7}{c}{image = Sawtooth} \\
\multicolumn{3}{c}{time: co=33s / gc=673s} & \multicolumn{4}{c}{energy: co=1,430,450 / gc=1,418,015} \\
\hline
& {\tiny ALL} &  {\tiny NON OCCL} &  {\tiny OCCL} &  {\tiny TEXTRD} & {\tiny TEXTRLS} & {\tiny D\_DISCNT }\\
Error &   1.46 &   0.61 &   7.92  &  0.63 &   0.33 &   1.56  \\
Bad Pixels &   3.93\%  & 1.35\% & 93.06\%  & 1.48\% &  0.14\%  & 5.96\% \\
\hline
Error &    1.49  &  0.70  &  7.88 &   0.73 &   0.40 &   1.60 \\
Bad Pixels &    3.99\% &  1.38\% & 94.02\%  & 1.49\%  & 0.31\% &  6.39\% \\
\hline
\end{tabular}

\vskip 0.3in

\begin{tabular}{ccccccc}
\hline
\multicolumn{7}{c}{image = Tsukuba} \\
\multicolumn{3}{c}{time: co=20s / gc=476s} & \multicolumn{4}{c}{energy: co=517,591 / gc=503,962} \\
\hline
& {\tiny ALL} &  {\tiny NON OCCL} &  {\tiny OCCL} &  {\tiny TEXTRD} & {\tiny TEXTRLS} & {\tiny D\_DISCNT }\\
Error &  1.30  &  0.99 &   5.41  &  1.00  &  0.97  &  2.01  \\
Bad Pixels &    4.77\% &  2.59\% & 87.38\% &  2.57\% &  2.61\% & 10.63\% \\
\hline
Error &   1.25  &  0.92 &   5.35 &   1.04  &  0.73  &  2.02\\
Bad Pixels &    4.24\% &  2.04\% & 87.60\% &  2.77\% &  1.05\% & 10.00\%\\
\hline
\end{tabular}

\vskip 0.3in

\begin{tabular}{ccccccc}
\hline
\multicolumn{7}{c}{image = Venus} \\
\multicolumn{3}{c}{time: co=35s / gc=573s } & \multicolumn{4}{c}{energy: co=1,253,764 / gc=1,246,078} \\
\hline
& {\tiny ALL} &  {\tiny NON OCCL} &  {\tiny OCCL} &  {\tiny TEXTRD} & {\tiny TEXTRLS} & {\tiny D\_DISCNT }\\
Error &    1.58  &  1.11 &   8.29 &   0.93  &  1.42  &  1.49  \\
Bad Pixels &    3.29\% &  1.65\% & 90.72\% &  1.38\%  & 2.20\%  & 7.28\% \\
\hline
Error &    1.47 &   0.95 &   8.33  &  0.81  &  1.18 &   1.31 \\
Bad Pixels &    3.58\%  & 1.93\% & 91.55\%  & 1.56\% &  2.68\% &  6.84\% \\
\hline
\end{tabular}
}
\end{center}

\section{Conclusions}

Cooperative optimization offers us a general, distributed optimization method for attacking hard optimization problems.
Soft decision making, message passing, and solution compromising are three important techniques 
	for achieving cooperation among agents in a cooperative optimization system.
The global optimality property of consensus solutions
	offers an appealing reason for agents in a system
	to compromise their solutions 
	so that conflicts in their solutions can be resolved.
The insights we gained at studying cooperative optimization
	might help us to apply the cooperation principle 
	to understand or solve more generic decision optimization problems arose from fields like 
	neurosciences, business management, political management, and social sciences.

\section{Appendix}

\subsection{Proof of Theorem~\ref{sufficient_condition}}

\begin{proof}
Although the proof of the theorem is simple and straightforward,
	the property it reveals is important for cooperative optimization.

Since ${\tilde x}$ is a consensus solution, substitute it into (\ref{cooperative_optimization}) we have
\[ \Psi^{(\infty)}_i ({\tilde x}_i) = (1-\lambda) E_i({\tilde x})  +
	\lambda \sum_{j}  w_{ij} \Psi^{(\infty)}_j({\tilde x}_j), \]
for $1 \le i \le n$. Sum them up, we have
\begin{eqnarray*}
\sum_i \Psi^{(\infty)}_i ({\tilde x}_i) & = & \sum_i \left( (1-\lambda) E_i({\tilde x})  +
	\lambda \sum_{j}  w_{ij} \Psi^{(\infty)}_j({\tilde x}_j) \right) \\
	& = & (1-\lambda) E({\tilde x})  +
	\lambda \sum_j \Psi^{(\infty)}_j({\tilde x}_j) \ .
\end{eqnarray*}
That is 
\begin{equation}
E({\tilde x}) = \sum_i \Psi^{(\infty)}_i ({\tilde x}_i) \ .
\label{t1a}
\end{equation}

For any $x$, from (\ref{cooperative_optimization}), we have
\[ \Psi^{(\infty)}_i (x_i) \le   (1 - \lambda) E_i(x)  +
	\lambda \sum_{j}  w_{ij} \Psi^{(\infty)}_j(x_j), \]
for $1 \le i \le n$. Sum them up, we have
\begin{eqnarray*}
\sum_i \Psi^{(\infty)}_i (x_i) & \le & \sum_i \left( (1 - \lambda) E_i(x)  +
	\lambda \sum_{j}  w_{ij} \Psi^{(\infty)}_j(x_j)  \right) \\
	& = & E(x)  +
	\lambda \sum_{j} \Psi^{(\infty)}_j(x_j) \ .
\end{eqnarray*}
That is 
\begin{equation}
E(x) \ge \sum_i \Psi^{(\infty)}_i (x_i) \ .
\label{t1b}
\end{equation}

Subtract (\ref{t1a}) from (\ref{t1b}), we have
\begin{equation}
E(x) - E({\tilde x}) \ge \sum_i \left( \Psi^{(\infty)}_i (x_i) - \sum_i \Psi^{(\infty)}_i ({\tilde x}_i) \right) \ . 
\label{t1c}
\end{equation}

Because
\[ \Psi^{(\infty)}_i (x_i) \ge \Psi^{(\infty)}_i ({\tilde x}_i) \ , \]
from (\ref{t1c}), we have
\[ E(x) - E({\tilde x}) \ge 0 \ . \]
Therefore, ${\tilde x}$ must be the global minimum of $E(x)$. 

This completes the proof. $\S$
\end{proof}

\subsection{Proof of Lemma~\ref{lemma_infinite_product}}
\label{appendix_lemma}

\begin{proof}
For any numbers $\lambda_1, \ldots, \lambda_k$ in $[0, 1)$, 
	the following inequality can be proved by the principle of mathematical induction,
\[ \lambda_1 \lambda_2 \ldots \lambda_k \ge 1 - (1-\lambda_1) - (1-\lambda_2) - \ldots - (1-\lambda_k) \ . \]
If $\sum^{\infty}_{k=1} (1-\lambda_k)$ converges,
	there exists $N$ such that for all $n \ge N$,
\[ (1-\lambda_N) + \ldots + (1-\lambda_n) < \frac{1}{2} \ . \]
Therefore, defining $g(n)$ as
\[ g(n) = \prod^{n}_{k=1} \lambda_k \ , \]
we have that for all $n \ge N$,
\[ \frac{g(n)}{g(N-1)} = \lambda_N \ldots \lambda_n \ge 1 - \left( (1-\lambda_N) + \ldots + (1- \lambda_n)\right) \ge \frac{1}{2} \ . \]
Therefore, the sequence $\{g(n)\}_{n \ge N}$ is a non increasing sequence bounded from below by 
	$g(N-1)/2 > 0$. 
It must have a positive limit $\epsilon > 0$ so that 
\[ \lim_{n \rightarrow \infty} \prod^{n}_{k=1} \lambda_k = \epsilon > 0 \ . \]

Using the inequality $1-x \le e^{-x}$ when $x \in [0,1)$, we have that
\[ g(n) = \prod^{n}_{k=1} (1 - (1-\lambda_k)) < e^{-\left( (1-\lambda_1) + \ldots + (1-\lambda_n) \right)} \ . \]
If $\sum^{\infty}_{k=1} (1-\lambda_k)$ is a divergent series, 
	we have
\[ \lim_{n \rightarrow \infty} \prod^{n}_{k=1} \lambda_k =0 \ . \]

This completes the proof. $\S$
\end{proof}


\end{document}